\documentclass[lettersize,journal]{IEEEtran}
\usepackage{graphicx}
\usepackage{epstopdf} 
\usepackage{amsmath,amsfonts}
\usepackage{array}
\usepackage[caption=false,font=normalsize,labelfont=sf,textfont=sf]{subfig}
\usepackage{textcomp}
\usepackage{stfloats}
\usepackage{url}
\usepackage{verbatim}
\usepackage{cite}

\usepackage[linesnumbered, ruled]{algorithm2e}
\SetKwRepeat{Do}{do}{while}%

\usepackage{float}
\usepackage[utf8]{inputenc} 
\usepackage[T1]{fontenc}    
\usepackage{hyperref}       
\usepackage{url}            
\usepackage{booktabs}       
\usepackage{amsfonts}       
\usepackage{nicefrac}       
\usepackage{microtype}      
\usepackage{xcolor}         
\usepackage{amsmath} 
\usepackage{multirow} 
\usepackage{makecell}
\usepackage{xcolor}
\usepackage{colortbl}
\usepackage{float}
\usepackage{wrapfig}
\usepackage{amsmath,amsfonts,amssymb,amsthm}
\usepackage{lipsum}       

\usepackage[capitalise]{cleveref} 
\crefformat{figure}{Fig.~#2#1#3} 
\crefformat{table}{Tab.~#2#1#3}  

\hypersetup{urlcolor=red, linkcolor=red}
\usepackage{tikz}

\definecolor{lime}{HTML}{A6CE39}
\DeclareRobustCommand{\orcidicon}{
	\begin{tikzpicture}
		\draw[lime, fill=lime] (0,0)
		circle[radius=0.16]
		node[white]{{\fontfamily{qag}\selectfont \tiny \.{I}D}};
	\end{tikzpicture}
	\hspace{-2mm}
}
\foreach \x in {A, ..., Z}{%
	\expandafter\xdef\csname orcid\x\endcsname{\noexpand\href{https://orcid.org/\csname orcidauthor\x\endcsname}{\noexpand\orcidicon}}
}

\begin{document}

\title{TSkel-Mamba: Temporal Dynamic Modeling via State Space Model for Human Skeleton-based Action Recognition}

\author{Yanan Liu\hspace{-1.5mm}\orcidA{},  Jun Liu\orcidD{}, Hao Zhang\orcidB{}, Dan Xu\orcidC{}, Hossein Rahmani\orcidE{}, Mohammed Bennamoun\orcidF{}, Qiuhong Ke\orcidG{}

\thanks{Yanan Liu (liuyanan@mail.ynu.edu.cn),  Hao Zhang and Dan Xu (danxu@ynu.edu.cn) are from the School of Information Science and Engineering, Yunnan University, Kunming, China.}

\thanks{Jun Liu (j.liu81@lancaster.ac.uk) and Hossein Rahmani are from Lancaster University, United Kingdom. }

\thanks{Mohammed Bennamoun (mohammed.bennamoun@uwa.edu.au) is with The University of Western Australia, Australia.}

\thanks{Qiuhong Ke (qiuhong.ke@monash.edu) is from Monash University, Melbourne, Australia.  }

}

\markboth{Journal of \LaTeX\ Class Files,~Vol.~14, No.~8, August~2021}%
{Shell \MakeLowercase{\textit{et al.}}: A Sample Article Using IEEEtran.cls for IEEE Journals}


\maketitle
\newcommand{\liu}[1]{\textcolor[rgb]{0.90, 0.0, 0.0}{#1}}
\begin{abstract}
Skeleton-based action recognition has garnered significant attention in the computer vision community. Inspired by the recent success of the selective state-space model (SSM) Mamba in modeling 1D temporal sequences, we propose TSkel-Mamba, a hybrid Transformer-Mamba framework that effectively captures both spatial and temporal dynamics. In particular, our approach leverages Spatial Transformer for spatial feature learning while utilizing Mamba for temporal modeling.
Mamba,  however,  employs separate SSM blocks for individual channels,  which inherently limits its ability to model inter-channel dependencies.      To better adapt  Mamba for skeleton data and  enhance Mamba`s ability to model temporal dependencies,  we introduce a Temporal Dynamic Modeling (TDM) block, which is a versatile plug-and-play component that integrates a novel Multi-scale Temporal Interaction (MTI) module.   The MTI module employs multi-scale Cycle operators to capture cross-channel temporal interactions, a critical factor in action recognition.
Extensive experiments on NTU-RGB+D 60, NTU-RGB+D 120, NW-UCLA and UAV-Human datasets demonstrate that TSkel-Mamba achieves state-of-the-art performance while maintaining  low inference time, making it both efficient and highly effective.
\end{abstract}

\begin{IEEEkeywords}
Action recognition, human skeleton, state space model, mamba, temporal dynamic.
\end{IEEEkeywords}

\section{Introduction}
\label{section1}
\IEEEPARstart{H}{uman} action recognition~\cite{Review1,Review2} is a key research area with wide applications in robotics~\cite{ISTA}, human-computer interaction~\cite{IGFormer}, and virtual reality~\cite{vr1,vr2}. Skeleton-based action recognition, in particular, has gained sustained attention in the computer vision community due to its robustness against background noise and disturbances caused by varying camera views~\cite{2s-AGCN,DGNN}.

To tackle skeleton-based action recognition, spatio-temporal architectures~\cite{ST-GCN, MS-G3D} have proven effective. However, most methods focus primarily on learning complex spatial patterns. These approaches use Graph Convolutional Networks (GCNs)~\cite{2s-AGCN, InfoGCN, STGCNplus, MS-G3D, ShiftGCN} to aggregate joint correlations based on the natural physical topology of the human body or employ spatial Transformers~\cite{IGFormer, ST-TR, HyperFormer} to model   joint dependencies within larger contextual windows. Since an action is composed of evolving poses over time~\cite{DPRL-GCN}, robust modeling of temporal dynamics is essential for further improving performance.

Although recent Convolutions~\cite{2s-AGCN,CTR-GCN} and Transformers~\cite{DSTA-Net,STST} showcase impressive performance, they face limitations: 1)  CNN-based methods, with their restricted local receptive fields, are not inherently designed to perceive long-range dependencies,  making it challenging to  capture robust temporal dynamics from complex motion
\begin{figure}[!]
  \centering
  \setlength{\abovecaptionskip}{-3pt}
  \includegraphics[width=\linewidth]{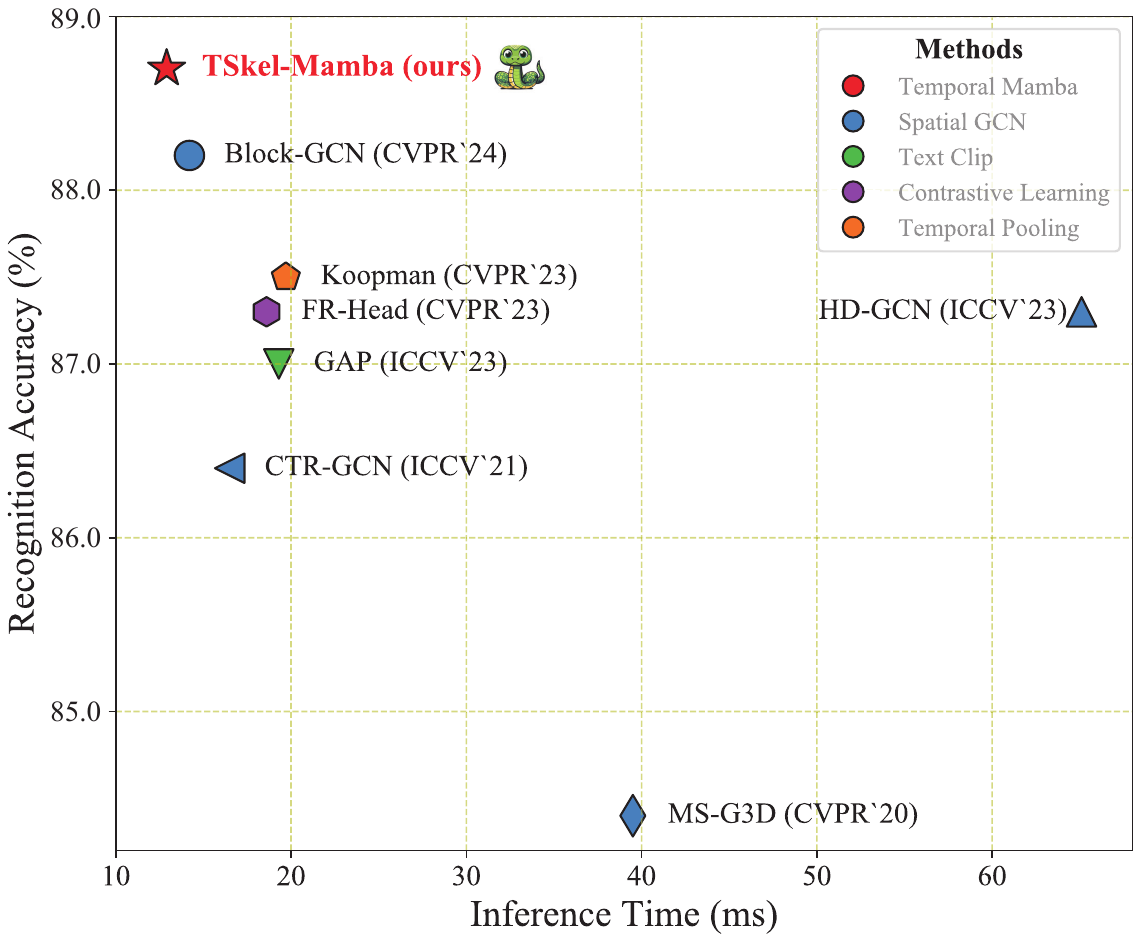}
  \caption{ Comparison of recognition accuracy and inference efficiency of our TSkel-Mamba against state-of-the-art methods on the NTU-RGB+D 120 Cross-Setup benchmark (joint modality). 
    Our Skel-Mamba achieves the best accuracy with the lowest inference cost.}
    \label{fig:IT}
\end{figure}

sequences.  2) The computational efficiency of Transformers shows limitations when processing long sequences.  Despite previous efforts~\cite{ST-TR} to design hybrid Transformer-CNN modules, their cost-effectiveness is still hampered by quadratic computational complexity with respect to the sequence length.   Therefore,  developing a 
efficient temporal block capable of learning long-term motion features for skeleton-based action recognition is a worthwhile endeavor.

  


Recently,  State-Space Models (SSMs)~\cite{SSM1,SSM2,SSM3} have shown a remarkable capability of sequence modeling, particularly excelling in efficiency of parallel training  with near-linear complexity.  The representative Mamba~\cite{Mamba} introduces a selection mechanism that stores flexible contexts through input-dependent SSM parameters, enabling adaptive modeling of long-range dependencies. Additionally, it also ensures efficient inference via hardware-aware algorithms~\cite{Mamba}, thereby demonstrating strong capacity for linear scaling in sequence modeling. Although recent studies~\cite{VMamba1,VMamba2} have extended Mamba to visual tasks involving images and videos, 
adapting Mamba for skeleton-based action recognition has not yet been explored.

Since the SSMs was originally designed for processing 1D sequences, 
{applying them to 3D skeleton data poses a challenge due to the complex spatial-temporal relationships between skeleton joints.} 
{To overcome this, we propose a simple yet effective hybrid Transformer-Mamba framework, \textbf{TSkel-Mamba}, for skeleton-based action recognition. In this framework, spatial and temporal information are learned separately—spatial patterns are captured by a Spatial Transformer, while temporal dynamics are modeled using a Mamba-based approach. This specialization allows the model to fully leverage the strengths of both architectures, leading to more effective action recognition.}

To improve temporal modeling, we introduce a Mamba-based Temporal Dynamics Modeling (TDM) block, which adopts a simple temporal scanning strategy to generate forward and backward sequences, allowing Mamba to effectively learn  temporal patterns. Since Mamba lacks the ability to model cross-channel temporal interactions~\cite{mambamixer}, which are crucial for understanding how different motion components evolve together in action recognition~\cite{MS-G3D, ShiftGCN}, we design a Multi-scale Temporal Interaction (MTI) module within TDM. The MTI module strengthens cross-channel temporal interactions by integrating features from adjacent frames at multiple scales, enriching motion representations and refining Mamba’s ability to model temporal dependencies.

The key contributions of this work are as follows:

\begin{itemize}
	\item We introduce TSkel-Mamba, a novel framework that explores Mamba's potential for modeling \textbf{T}emporal information in \textbf{Skel}eton sequences for action recognition, making it a pioneering attempt  of Mamba in this domain.
	\item 
    We introduce a Temporal Dynamics Modeling (TDM) block, which includes a Multi-scale Temporal Interaction (MTI) module to enhances Mamba’s ability to model temporal dependencies by improving cross-channel temporal interactions. 
	\item 
    Our TSkel-Mamba achieves state-of-the-art performance on four challenging benchmark datasets while maintaining efficient inference (as shown in \cref{fig:IT}), demonstrating its effectiveness in both accuracy and computational cost.
\end{itemize}

\section{Related Work}

\subsection{Skeleton-based Action Recognition}

Early research utilized Recurrent Neural Networks (RNNs)~\cite{RNN1,RNN2,RNN3}  to capture temporal dependencies, but their poor ability to learn spatial patterns led to suboptimal performance. Convolutional Neural Networks (CNNs))~\cite{CNN1,VA-CNN,CNN2} combined with pseudo-images also fail to  effectively capture spatial interactions. Recently,  a  spatio-temporal graph representation~\cite{ST-GCN} utilizing Graph Convolutional operators has gained significant attention, prompting a shift towards learning spatial topologies. However, temporal modeling also play a crucial role in skeleton-based action recognition.

\textbf{Efforts in Spatial Learning.}  Early work~\cite{ST-GCN} introduced fixed adjacency matrices with physical topology embeddings for GCNs, achieving milestone performance.  However, this restricts long-range spatial interactions among joints, which may hold substantial semantic relevance despite their non-adjacency.  Considering the complexity of multi-joint coordination, most recent methods~\cite{2s-AGCN,CTR-GCN,MS-G3D,InfoGCN} explore learnable topology for extracting multi-scale spatial dependencies. And some methods~\cite{DSTA-Net,ST-TR,IGFormer,ISTA} employ Transformers for global joint interactions. However, learnable topology may overly forget inductive biases from physical topology during training, and Transformers overlook them entirely.  Therefore, state-of-the-art methods~\cite{BlockGCN,HyperFormer} focus on finding a "threshold" to assess spatial interactions, determining what to select and discard.


\textbf{Efforts in Temporal  Learning.} Recent studies~\cite{DPRL-GCN,2s-AGCN,TCA-GCN} focused on  more valuable temporal information or temporal pooling to improve recognition performance.  \cite{DPRL-GCN} selected key frames using  deep progressive reinforcement learning.    \cite{Koopman} propose  a parameterized high-order Koopman pooling. Alternatively, prior studies~\cite{CTR-GCN,BlockGCN,HD-GCN,InfoGCN,FGCN} employed temporal Convolution to aid heavy spatial-dominated networks,  but it is not robust in handling complex temporal interactions. While several approaches~\cite{DSTA-Net,ST-TR} introduced temporal Transformer, their computational costs outweigh the performance gains.  We noted that the \textsl{spatial-temporal} network architecture is applied in most advanced works~\cite{ST-GCN,2s-AGCN,CTR-GCN,HD-GCN,HyperFormer}. However,  temporal feature operator is not fully developed. In reviewing RNNs, although they are suitable for causal sequence, suffering from gradient explosion and limited state space~\cite{Mamba,SSM2}.  Reflecting on the development of spatial module, we seek to design a module that retains valuable interactions and discards redundant temporal information.

\subsection{Mamba in Computer Vision}

Mamba~\cite{Mamba} is a selective State-Space Model (SSM) that offers modeling capabilities comparable to Transformers while ensuring near-linear scalability with respect to sequence length, sparking broad exploration in  CV community~\cite{Mamba1,Mamba2,Mamba3,Mamba4}.  \cite{VMamba0} introduced a  2D selection strategy to adapt Mamba from 1D sequences to grid-based 2D images. \cite{PointMamba} developed PointMamba, a framework for point cloud analysis that captures inter-group relationships with Mamba. \cite{MotionMamba} presented Motion Mamba, a SSM based motion generation model,  composed of the \textit{UNet}-structure and a hierarchical mamba block. Therefore, Mamba is a highly suitable solution with the potential to be developed into a pioneering temporal plugin that effectively improve  performance for skeleton-based action recognition. 

\vspace{-5pt}

\section{Method}
\label{Sec3}
\vspace{-5pt}
\subsection{Preliminaries}
\textbf{Selective State-Space Model (SSM).} Selective SSM~\cite{Mamba} represents a dynamic system by the state at time step $t$. The input sequence $x(t)\in\mathbb{R}^{D}$ is mapped to the output sequence $y(t)\in\mathbb{R}^{D}$ via a  hidden state $h(t)\in\mathbb{R}^{N}$, with the evolution of the state governed by the Eq.\eqref{eq1}:

\begin{equation}
    \begin{aligned}
\label{eq1}
&h'{(t)}=\boldsymbol{A}h{(t)}+\boldsymbol{B}x{(t)},\\
    &y{(t)}=\boldsymbol{C}h{(t)},
    \end{aligned}
\end{equation}
  
    

where $\boldsymbol{A}\in\mathbb{R}^{N\times N}$ is the learnable evolution matrix. $\boldsymbol{B}\in\mathbb{R}^{N\times D}$ and $\boldsymbol{C}\in\mathbb{R}^{D\times N}$ are two learnable projection matrices. 
$h'{(t)}$ denotes the derivative of $h{(t)}$.  In fact, the SSM needs to be discretized using a step size $\Delta$ to transform the continuous parameters $\boldsymbol{A},\boldsymbol{B}$ to discrete parameters $\boldsymbol{\Bar{A}},\boldsymbol{\Bar{B}}$, namely $\boldsymbol{\Bar{A}}=exp(\Delta\boldsymbol{A})$ and $\boldsymbol{\Bar{B}}=(\Delta\boldsymbol{A})^{-1}(exp(\Delta\boldsymbol{A})-\boldsymbol{I})\cdot \Delta\boldsymbol{B}$. The Eq.\eqref{eq1} is transformed into Eq.\eqref{eq2}.
\begin{equation}
        \begin{aligned}
	\label{eq2}
	&h_{t}=\boldsymbol{\Bar{A}}h_{t-1}+\boldsymbol{\Bar{B}}x_{t},\\
        &y_{t}=\boldsymbol{C}h_{t},
        \end{aligned}
    \end{equation}
However,  SSM exhibits numerical instability and high computational costs. To address these limitations, the structured SSM (S4~\cite{SSM3}) introduces the \textit{HIPPO}~\cite{Hippo} matrix. Notably, Mamba employs hardware-aware algorithms to facilitate parallel computation of dynamic parameter matrices.

\textbf{Skeleton Action Representation.} 
 Given a skeleton data $p\in \mathbb{R}^{B\times C\times N}$, 
consisting of feature of $N$  human joints $V_n\in \mathbb{R}^{C} ,n\in [1,N]$, where $N$ is the number of  joints, {$B$ is the batch size} and  $C$ is the number of channels,  an action can be viewed as a temporal sequence $\boldsymbol{P}=\{p_1\dots p_T\}\in \mathbb{R}^{B\times C\times T\times N}$, consisting of $T$ frames of skeletons, 
The initial skeleton representation can be represented as $\boldsymbol{H}_{(0)}=\boldsymbol{P}\in \mathbb{R}^{B\times C_{in}\times T \times N}$, where $C_{in}$ is the  number of input channels, initialized to 3, representing the 3D coordinates $(x_n,y_n,z_n)$ in Euclidean space.  





\begin{figure*}[tb]
	\centering 

	\includegraphics[width=1\textwidth]{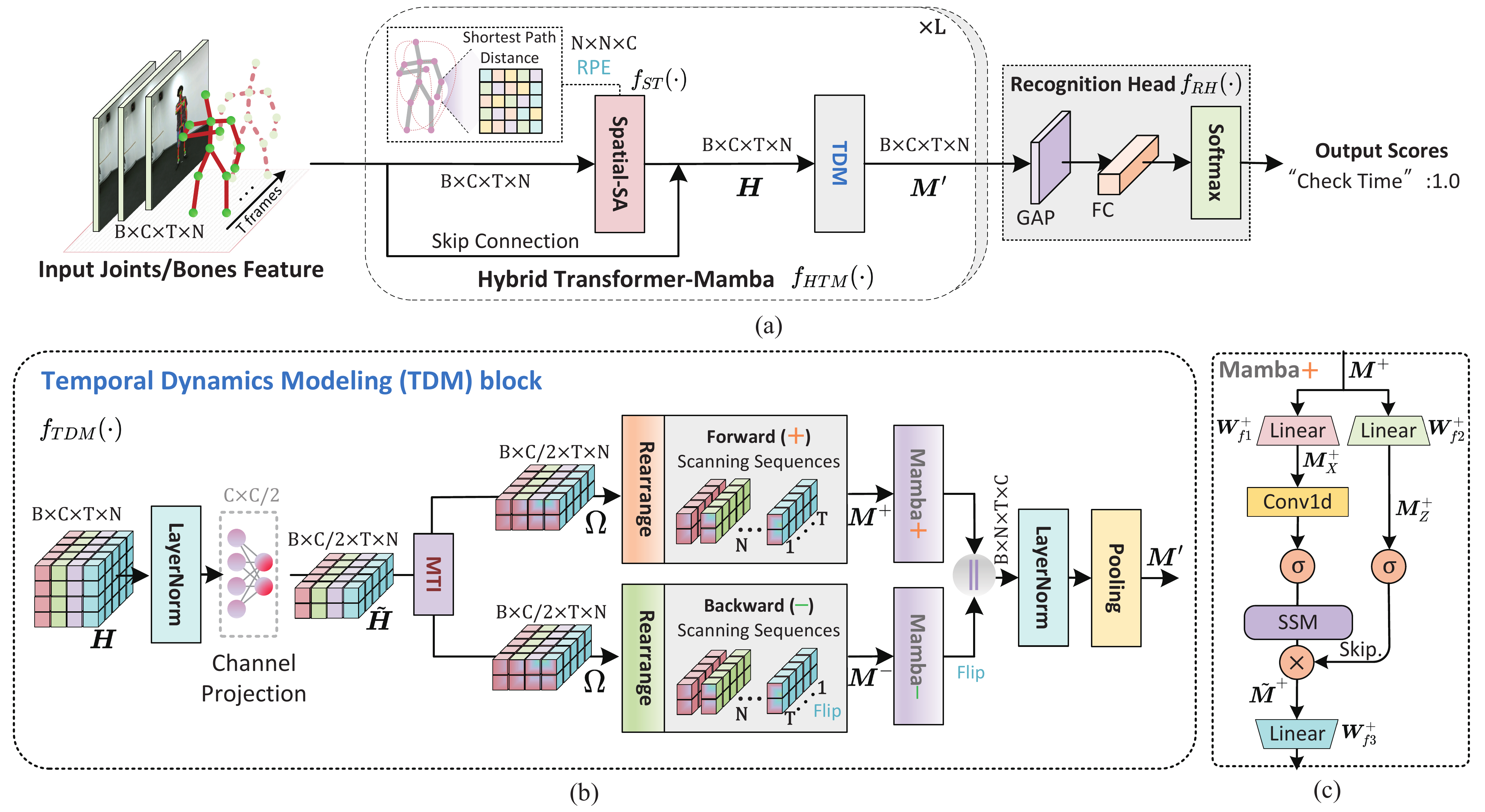}
	\caption{\textbf{(a)} The pipeline of our TSkel-Mamba. Spatial-SA denotes  spatial Transformer with the \textsl{relative position encoding} (RPE) calculated by  joint \textit{shortest path distance}. \textbf{(b)} Overall architecture of proposed Temporal Dynamic Modeling (TDM) block. MTI denotes Multi-scale Temporal Interaction module. Mamba denotes the selective State-Space Model mamba~\cite{Mamba}. $||$ denotes the concatenate operation.   {\textbf{(c)} Diagram of the structure of Mamba.  (Best viewed in color)}}

	\label{fig:OA}
	
\end{figure*}

\subsection{Overall Architecture}
As illustrated in  \cref{fig:OA}(a), the proposed TSkel-Mamba $\boldsymbol{F}(\cdot)$ comprises $L$ layers of  a Hybrid Transformer-Mamba (HTM) feature extractor $\{f_{HTM}^{\left(l\right)}\}^{L}_{l=1}$, which maintains a
common \textsl{spatial-temporal} architecture. Additionally, it includes a recognition head $f_{RH}(\cdot)$, consisting of Global Average Pooling and a Fully Connected layer, which maps the extracted features to action classes. 
In particular, each HTM layer consists of: (1) 
A Spatial Transformer block $f_{ST}(\cdot)$, which leverages multi-head self-attention \cite{AAN} and relative position encoding \cite{HyperFormer} to effectively model long-range spatial dependencies. 
(2) 
A carefully designed Mamba-based Temporal Dynamics Modeling (TDM) block $f_{TDM}(\cdot)$ (see \cref{fig:OA}(b)), which incorporates specialized mechanisms tailored for skeleton data, enabling effective learning of complex temporal information. Finally, the extracted high-level features are processed by $f_{RH}(\cdot)$ to generate the final action classification scores. 
{Below, we describe the details.}

\subsection{Mamba-based Temporal Dynamics Modeling}
\label{subsec33}
The overall architecture of the proposed TDM block is illustrated in \cref{fig:OA}(b). Given a skeleton feature 
produced by the spatial Transformer, we first apply channel projection to downsample the channel dimension, reducing the computational cost of subsequent modules. Since directly feeding the features into Mamba for temporal modeling does not effectively capture cross-channel temporal interactions--which are essential for understanding how different motion components evolve together and achieving a more comprehensive representation of action dynamics for better   recognition~\cite{MS-G3D, ShiftGCN, CTR-GCN}--we introduce a Multi-scale Temporal Interaction (MTI) module. The MTI module enhances cross-channel temporal interactions by aggregating features from adjacent frames at multiple scales, leading to richer and more informative motion representations. Inspired by VideoMamba~\cite{VideoMamba}, which employs Bidirectional SSM~\cite{VideoMamba} for video action recognition, we further generate forward and backward scanning sequences based on a pure temporal scanning strategy, enabling Mamba to effectively capture global temporal dependencies and improve overall sequence understanding. Below, we provide a detailed description of each component.

\textbf{Channel Projection:}  As shown in \cref{fig:OA}(b), given the skeleton feature   $\boldsymbol{H}\in \mathbb{R}^{B\times C\times T\times N}$ output by spatial Transformer,  
where $B$ is batch size, $C$ is the number of channels, $T$ is the number of frames and $N$ is the number of spatial joints, we first apply layer normalization $LN(\cdot)$. Then we perform channel projection via a $1\times 1$ convolution, followed by a BatchNorm layer and a ReLU activation function,   yielding an intermediate representation $\Tilde{\boldsymbol{H}}\in \mathbb{R}^{B\times C/2 \times T\times N}$. With this  process, we reduce the parameter burden of TDM block, where we process the features in two streams. 
{In theory, directly processing the original $\boldsymbol{H} \in \mathbb{R}^{B\times C\times T\times N}$ with a linear layer would require a weight matrix of size $\mathbb{R}^{C\times C}$ to generate a feature map with $C$ channels, resulting in $C^2$ parameters. However, by processing two $\Tilde{\boldsymbol{H}}$ representations with $C/2$ channels each and then concatenating the output features to produce a final feature map with $C$ channels, only $2 \times (C/2)^2$ parameters are required. This approach significantly reduces computational overhead while maintaining the capacity to model multi-channel information effectively. 
}

\begin{figure*}[!]
	\centering 
	
	\includegraphics[width=1\textwidth]{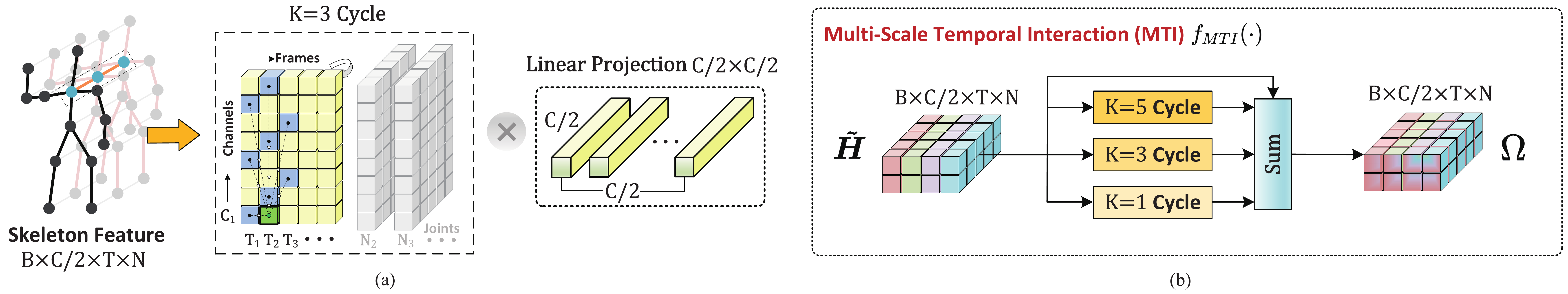}
	\caption{{\textbf{(a)} The diagram of Cycle Fully Connection {(FC)} layers to enhance 
    {cross-channel temporal}
    interaction.   $K$ denotes {the 
    kernel size, which is set to 3}. 
    As the channel increases, it {cycles 
    along} the temporal dimension with a step size of $\{-1,0,1\}$. \textbf{(b)}  The diagram of Multi-scale Temporal Interaction Module. 
    $K=1$, $K=3$ and $K=5$ denote the kernel sizes with different scales.} }
	\label{fig:Cycle}
	
\end{figure*}

{\textbf{Multi-scale Temporal Interaction Module:}} 
{As shown in {\cref{fig:Cycle}(b)},   $\Tilde{\boldsymbol{H}}\in \mathbb{R}^{B\times C/2 \times T\times N}$ is then processed with our  MTI module  $f_{MTI}(\cdot)$, which  is built on Cycle Fully Connected (FC)  layer~\cite{CycleMLP} $f_{Cycle}(\cdot)$.} 
For a certain joint,   Cycle FC is designed to aggregate cross-channel features within its temporal {adjacent frames}, followed by a linear projection. Specifically, let {$\Tilde{\boldsymbol{H}}(c,t,n)$}  be the $c$-th channel feature of the $n$-th spatial joint in the $t$-th frame, {the    $f_{Cycle}$ operation} is formulated as:
\begin{equation}
\label{eq4}
    {f_{Cycle}^{K}(\Tilde{\boldsymbol{H}}_{(:,t,n)})=  \sum^{C_{in}}_{c=0}{\Tilde{\boldsymbol{H}}_{(c,t+\delta_t(c),n)}}\cdot \boldsymbol{W}_{c}+\boldsymbol{b},}
\end{equation}

\noindent where {$C_{in}$ {denotes the number of input channels}.} $\delta_t(c)=(c\mod K)-1$ is the time offset and 
{$K$ is the kernel size}. $\boldsymbol{W}_{c}\in \mathbb{R}^{C_{in} \times C_{out}}$ and $\boldsymbol{b}\in\mathbb{R}^{C_{out}}$ are weight matrix and bias of the linear layer. $C_{out}$ {denotes the number of output channels}.  In \cref{fig:Cycle}(a), take the kernel size $K=3$ as an example, as the channel increases, the interaction window {cycles 
along the temporal dimension with a step size of $\{-1,0,1\}$.} 
{With this approach,} Cycle FC   effectively facilitates {cross-channel temporal interactions.} 
{Considering that temporal dynamics of human actions often span across varying time scales, we}
select several Cycle FC layers with different temporal kernel sizes to further construct a Multi-Scale {Temporal Interaction} (MTI) module, as shown in {\cref{fig:Cycle}(b)}. This {enables the  model to simultaneously capture channel interaction at diverse temporal scales, providing a more robust and comprehensive representation of the temporal evolution within the data. Formally,}
\begin{equation}
\label{eq5}
    {f_{MTI}(\Tilde{\boldsymbol{H}})=\Tilde{\boldsymbol{H}}+\sum^{k\in S_{K}}_{K}{f_{Cycle}^{K}(\Tilde{\boldsymbol{H})}},}
\end{equation}
where  $S_{K}$ denotes a  set of  kernel sizes at different scales.  

{{\textbf{Scanning Sequence  Generation:}}}  With the enhanced skeleton feature $ {\boldsymbol{\Omega}}\in \mathbb{R}^{B\times C/2\times T\times N}$ output by the MTI,  a key challenge in using {Mamba} 
for sequence   modeling  is constructing an effective scanning sequence that preserves temporal dynamics. Since {Mamba}~\cite{Mamba} is designed for 1D sequences, it lacks inherent compatibility with 3D skeleton data, which has complex spatio-temporal structures. A straightforward approach is to flatten joints into a 1D sequence using either: {(1) \textit{temporal-spatial scanning}, where joints are first ordered temporally and then spatially as ${v_{1,1},\dots, v_{1,T},\dots, v_{N,1},\dots, v_{N,T}}\in \mathbb{R}^{{B}\times TN\times {C/2}}$, or (2) \textit{spatial-temporal scanning}, where joints are first ordered spatially and then temporally as ${v_{1,1},\dots, v_{N,1},\dots, v_{1,T},\dots, v_{N,T}}\in \mathbb{R}^{{B}\times NT\times {C/2}}$. However, these methods disrupt temporal continuity—e.g., the $N_{th}$ joint in frame $t$ becomes adjacent to the $1_{st}$ joint in frame $(t+1)$, despite lacking direct temporal correlation—leading to weak performance (\cref{tab3}).}

{
{Inspired by spatial-temporal specialization in the overall framework,  i.e., spatial and temporal features are targeted for specialized learning by different experts, we treat {Mamba} as a temporal expert, focusing solely on capturing temporal dependencies. Thus, we adopt a pure \textit{temporal scanning} strategy, generating  a {temporal token sequence ${v_{1},\dots, v_{T}}_{(n)}\in \mathbb{R}^{T\times C}$ for each joint $n\in[1,N]$. {Given the input  $ {\boldsymbol{\Omega}}\in \mathbb{R}^{B\times C/2\times T\times N}$, there will be {$BN$} temporal token sequences.}}

\textbf{Bidirectional Temporal Dynamic Modeling:} 
With the scanning strategy mentioned above,  we perform forward 
scanning of the enhanced features  {$\boldsymbol{\Omega}$} for Mamba to process. 
Denote the forward ($+$) 1D  scanning sequences as {$\boldsymbol{M}^{+}\in\mathbb{R}^{BN\times T\times C/2}$}. 
We then flip {$\boldsymbol{M}^{+}$} along the temporal dimension to generate    the backward ($-$) scanning sequence {$\boldsymbol{M}^{-} \in\mathbb{R}^{BN\times T\times C/2}$}. 
Next, the 
{intermediate} $C'$-dimension features  {$\boldsymbol{M}_{X}^{+},\boldsymbol{M}_{X}^{-}\in\mathbb{R}^{BN\times T\times C'}$} and {$\boldsymbol{M}_{Z}^{+},\boldsymbol{M}_{Z}^{-}\in\mathbb{R}^{BN\times T\times C'}$} are created by different linear projections {$\boldsymbol{W}^{+}_{f1},\boldsymbol{W}^{+}_{f2},\boldsymbol{W}^{-}_{f1},\boldsymbol{W}^{-}_{f2}\in\mathbb{R}^{C/2\times C'}$. \cref{fig:OA}(c) illustrates the process of generating the forward  sequence. The backward scanning process follows the same procedure, except that it uses different weights and produces distinct intermediate representations. Formally:}
{\begin{equation}
\begin{aligned}
	\label{eq6}
	&\boldsymbol{M}_{X}^{+}=\boldsymbol{M}^{+}\boldsymbol{W}^{+}_{f1}, \boldsymbol{M}_{Z}^{+}=\boldsymbol{M}^{+}\boldsymbol{W}^{+}_{f2},\\
        &\boldsymbol{M}_{X}^{-}=\boldsymbol{M}^{-}\boldsymbol{W}^{-}_{f1}, \boldsymbol{M}_{Z}^{-}=\boldsymbol{M}^{-}\boldsymbol{W}^{-}_{f2},
\end{aligned}
\end{equation}}

\noindent Then the dual-stream branches are designed to process features in both forward  and backward  order:
{\begin{equation}
\begin{aligned}
	\label{eq7}
	&\Tilde{\boldsymbol{M}}^{+}=SSM^{+}(\sigma(Conv1D^{+}(\boldsymbol{M}_{X}^{+})))\odot \sigma(\boldsymbol{M}_{Z}^{+}),\\
        &\Tilde{\boldsymbol{M}}^{-}=SSM^{-}(\sigma(Conv1D^{-}(\boldsymbol{M}_{X}^{-})))\odot \sigma(\boldsymbol{M}_{Z}^{-}),
\end{aligned}
\end{equation}}

\noindent where $SSM(\cdot)$ denotes the selective SSM~\cite{Mamba} for efficiently and robustly capturing temporal dynamics. {$\Tilde{\boldsymbol{M}}^{+},\Tilde{\boldsymbol{M}}^{-}\in\mathbb{R}^{BN\times T\times C'}$} denote the intermediate representation.
$\sigma(\cdot)$ is the activation function. $Conv1D(\cdot)$ denotes the 1D Convolution, which is suitable for processing causal temporal sequence.  $\odot$ is the \textit{Hadamard Product}, mapping features to high-dimensional space to enhance nonlinearity. 
Finally, the enriched bidirectional features, after projection, are fused using a concatenation operation $Cat(\cdot)$:
{\begin{equation}
    \label{eq8}
{\boldsymbol{M}'=Pool((LN(Cat(\Tilde{\boldsymbol{M}}^{+}\boldsymbol{W}^{+}_{f3},Flip(\Tilde{\boldsymbol{M}}^{-}\boldsymbol{W}^{-}_{f3})))))},
\end{equation}}

\noindent  where  $Flip$ denotes the  flipping operation {along the temporal dimension}. $\boldsymbol{W}^{+}_{f3},\boldsymbol{W}^{-}_{f3}\in\mathbb{R}^{C'\times C/2}$ denote the output projection weights. $LN(\cdot)$ denotes the layer normalization.  {{$Pool(\cdot)$ denote the temporal pooling layer for temporal downsampling}}. $\boldsymbol{M}'$ denotes the final output of the TDM block. 

\textbf{Pseudo-code of TDM:} We provide a pseudo-code for TDM to improve methodological clarity and facilitate comprehension and reproduction, as shown in Algorithm~\ref{aa1}.

\begin{algorithm}[h]
      \label{aa1}
  \KwIn{skeleton-based action representations $\boldsymbol{H}:\textcolor{olive}{(B,C,T,N)}$}
  \KwOut{temporally-enhanced representations $\boldsymbol{M}':\textcolor{olive}{(B,C,T',N)}$}
  \textcolor{gray}{\tcp{normalize the input action representations $\boldsymbol{H}$}}
  $\boldsymbol{H}:\textcolor{olive}{(B,C,T,N)}:\leftarrow Permute(Norm(Permute(\boldsymbol{H})))$\;
  \textcolor{gray}{\tcc{perform channel projection to $C/2$ to reduce parameter burden}}
  $\Tilde{\boldsymbol{H}}:\textcolor{olive}{(B,C/2,T,N)}:\leftarrow Conv_{1\times1}(\boldsymbol{H})$\;
  \textcolor{gray}{\tcc{employ  MTI module to enhance cross-channel temporal interactions}}
  \For{$K$ in $S_{K}$} {
  $\Tilde{\boldsymbol{H}}=\Tilde{\boldsymbol{H}}+f_{Cycle}^{K}(\Tilde{\boldsymbol{H}})$\;
  }
  \textcolor{gray}{\tcp{the enhanced representations $\boldsymbol{\Omega}$}}
  $\boldsymbol{\Omega}:\textcolor{olive}{(B,C/2,T,N)}\leftarrow \Tilde{\boldsymbol{H}}$\;
  \textcolor{gray}{\tcc{construct  forward ($+$), backward ($-$) scanning sequence $\boldsymbol{M}^{+}$,$\boldsymbol{M}^{-}$}}
  $\boldsymbol{M}^{+}:\textcolor{olive}{(BN,T,C/2)}\leftarrow Reshape(\boldsymbol{\Omega})$\;
  $\boldsymbol{M}^{-}:\textcolor{olive}{(BN,T,C/2)}\leftarrow Flip(\boldsymbol{M}^{+})$\;
  \textcolor{gray}{\tcc{model bidirectional temporal dynamic }}
  \For{$i$ in $\{+,-\}$} {
    $\boldsymbol{M}^{i}_{X}:\textcolor{olive}{(BN,T,C')}\leftarrow Linear^{i}_{f1}(\boldsymbol{M}^{i})$\;
    $\boldsymbol{M}^{i}_{Z}:\textcolor{olive}{(BN,T,C')}\leftarrow Linear^{i}_{f2}(\boldsymbol{M}^{i})$\;
    \textcolor{gray}{\tcp{perform the bidirectional SSM}}
    $\Tilde{\boldsymbol{M}}^{i}:\textcolor{olive}{(BN,T,C')}=SSM^{i}(SiLU(Conv1D^{i}(\boldsymbol{M}_{X}^{i})))$\;
    \textcolor{gray}{\tcp{gated mechanism}}
    $\Tilde{\boldsymbol{M}}^{i}:\textcolor{olive}{(BN,T,C/2)}=Linear^{i}_{f3}(\Tilde{\boldsymbol{M}}^{i}\odot SiLU(\boldsymbol{M}_{Z}^{i}))$\;
  }
  \textcolor{gray}{\tcp{bidirectional feature fusion}}
    $\Tilde{\boldsymbol{M}}:\textcolor{olive}{(B,C,T,N)}=Reshape(Norm(Cat(\Tilde{\boldsymbol{M}}^{+},Flip(\Tilde{\boldsymbol{M}}^{-}))))$\;
  \textcolor{gray}{\tcc{temporal pooling to support temporal downsampling}}
    $\boldsymbol{M}':\textcolor{olive}{(B,C,T',N)}=TPooling(\Tilde{\boldsymbol{M}})$\;
    \textbf{Return} $\boldsymbol{M}'$
  \caption{Mamba-based Temporal Dynamics Modeling (TDM) Block}
\end{algorithm}

\subsection{Spatial Transformer with Topological Positional Encoding}
Despite primarily focusing on temporal framework design, exploring a compatible spatial expert for \textit{spatio-temporal} architecture is also essential. As shown in  \cref{fig:OA}(a), this work opts for a basic spatial Transformer~\cite{ST-TR}, introducing a  \textit{relative position encoding}  inspired by  \cite{RPE,HyperFormer} to alleviate the topological forgetting issue~\cite{BlockGCN}:
\begin{equation}
	\label{eq9}
    \boldsymbol{H}_{SA}=softmax(\boldsymbol{Q}\cdot \boldsymbol{K}^\intercal+\boldsymbol{Q}\cdot \boldsymbol{R}^\intercal)\cdot \boldsymbol{V},
\end{equation}
where $\boldsymbol{Q},\boldsymbol{K},\boldsymbol{V}\in\mathbb{R}^{T\times N\times  C}$ are different projections of the input  tensor. $\boldsymbol{R}\in\mathbb{R}^{N\times N\times C}$ denotes a parameterized matrix based on  joint \textit{shortest path distance}~\cite{HyperFormer} in skeleton topology.  This is a simplified formula that omits \textit{multi-head attention}. Our TDM shows excellent compatibility with this  spatial module, resulting in significant performance improvements.  Finally, the integrated Hybrid backbone, TSkel-Mamba, achieves  state-of-the-art performance for action recognition.

\subsection{Covariance Pooling with Knowledge distillation}
Next, we introduced a supporting technique via \textbf{C}ovariance \textbf{P}ooling\cite{CP} with \textbf{K}nowledge \textbf{D}istillation\cite{DKD} (\textbf{CPKD}) to highlight the scalability and deployment potential of our proposed TSkel-Mamba architecture.  Covariance Pooling (CP) has proven to be an effective alternative to the Global Average Pooling (GAP) in video action recognition~\cite{CP}, replacing GAP with CP in our TSkel-Mamba can improve performance. However, this substitution introduces a non-negligible number of parameters. To address this, we propose an efficient solution called CPKD (\cref{tab1}), which employs logit knowledge distillation~\cite{DKD} to boost performance without increasing the parameter count. Specifically, we use a pretrained TSkel-Mamba with CP as the teacher and a TSkel-Mamba with GAP as the student for distillation, enabling the model to achieve further performance gains while maintaining the model parameters.

This section elaborates on the methodology details of CPKD. Covariance pooling captures second-order motion information of human action features by computing the covariance matrix across feature channels, thereby preserving richer spatiotemporal and cross-channel interactions. This capability is particularly critical for recognizing complex actions and has been successfully applied in video-based action recognition~\cite{Gao_2019_CVPR}. Therefore, this motivates us to replace the original global average pooling (GAP, as shown in  Fig.~\textcolor{red}{\ref{fig:OA}}(a)) with covariance pooling (CP) to achieve better action recognition. Specifically,  given the final output $\boldsymbol{M}'\in\mathbb{R}^{B\times C\times T\times N}$ of our proposed TDM block, it is reshaped into skeleton feature  $\boldsymbol{O}\in\mathbb{R}^{B\times C\times d}$, where $d=TN$. The steps are as follows:

\textbf{1)} We perform covariance pooling-an operation computing a second-order covariance matrix as:
\begin{equation}
\label{eq1}
\boldsymbol{\Sigma}=\boldsymbol{O}\tilde{\boldsymbol{I}}\boldsymbol{O}^\top, \tilde{\boldsymbol{I}}=\frac{1}{d}(\boldsymbol{I}-\frac{1}{d}\boldsymbol{I}_{1}),\\    
\end{equation}
where $\boldsymbol{I}\in\mathbb{R}^{d\times d}$ and $\boldsymbol{I}_1\in\mathbb{R}^{d\times d}$ are identity matrix and all-ones matrix. 

\textbf{2)} We perform matrix square root normalization on the covariance matrix $\boldsymbol{\Sigma}$ to enhance feature discriminability. Generally, the covariance matrix is a symmetric positive (semi-)definite (SPD) matrix~\cite{CP}, which implies the existence of its matrix square root. For example, guided by eigenvalue decomposition,  $\boldsymbol{\Sigma}$ has a square root $\boldsymbol{Y}=\boldsymbol{U}diag(\lambda_i^{1/2})\boldsymbol{U}^\top$, where $diag(\lambda_i^{1/2})$ is a diagonal matrix formed by the eigenvalues $\lambda_i$ of $\boldsymbol{\Sigma}$, and $\boldsymbol{U}$ is an orthogonal matrix.  However, its implementation on GPU remains challenging. Following prior work in video recognition~\cite{Gao_2019_CVPR}, we also compute $\boldsymbol{Y}$ via \textsl{Newton-Schulz Iteration}~\cite{Levesley_2009}, the simplified formulation can be expressed as:
\begin{equation}
\begin{aligned}
	\label{eq2}
	&\boldsymbol{Y}_{k+1}=\frac{1}{2}\boldsymbol{Y}_{k}(3\boldsymbol{I}-\boldsymbol{Z}_{k}\boldsymbol{Y}_{k}),\\
        &\boldsymbol{Z}_{k+1}=\frac{1}{2}(3\boldsymbol{I}-\boldsymbol{Z}_{k}\boldsymbol{Y}_{k})\boldsymbol{Z}_{k},
\end{aligned}
\end{equation}
where $\boldsymbol{Z}_{0}=\boldsymbol{I}$ and $\boldsymbol{Y}_{0}=\boldsymbol{\Sigma}/tr(\boldsymbol{\Sigma})$ , $tr(\boldsymbol{\Sigma})=\Sigma_{i}\lambda_i$ is a trace-based scaling factor for normalization. $\boldsymbol{Y}_{k},\boldsymbol{Z}_{k},k\in[0,K-1]$ denote the intermediate variables in the $k$-th iteration. Finally, the square root matrix $\boldsymbol{Y}=\sqrt{tr(\boldsymbol{\Sigma})}\boldsymbol{Y}_{K}$ needs to be rescaled via $tr(\boldsymbol{\Sigma})$ to restore the original data magnitude. 

\textbf{3)} We perform upper-triangular vectorization--extracting the upper triangular portion of symmetric matrices ($\boldsymbol{Y}$) into vectors to reduce redundancy and lower dimensionality. Ultimately, high-dimensional motion features are processed through FC layers and a Softmax layer to generate action class logit.

However, this simple trick introduces additional computational costs ($e.g.$, the parameter count in FC layers increases from $C\times D$ to $C(C+1)/2\times D$, where C denotes the number of the last channels and $D$ denotes the number of action classes), particularly during inference, which hinders the expected optimization benefits. To overcome this, we introduce a decoupled knowledge distillation~\cite{DKD}, which is a parameter-free logit-based distillation scheme. Specifically, we use a pretrained TSkel-Mamba with CP as the teacher to generate the predicted probabilities $P^{\mathrm{T}}$   and a TSkel-Mamba with GAP as the student to generate the predicted probabilities $P^{\mathrm{S}}$  for knowledge distillation. The formulation can be expressed as:
\begin{eqnarray}
    \label{eq3}	\boldsymbol{L}_{KD}\left(P^{\mathrm{S}},P^{\mathrm{T}}\right)=\alpha\underbrace{\mathrm{KL}\left(P_b^{\mathrm{S}} \Vert P_b^{\mathrm{T}}\right)}_{\mathcal{L}_{TAKD}} +\beta\underbrace{\mathrm{KL}\left(P_m^{\mathrm{S}} \Vert P_m^{\mathrm{T}}\right)}_{\mathcal{L}_{NAKD}},
\end{eqnarray}
where $\alpha$ and $\beta$ are two hyper-parameters ($\alpha$=1 and $\beta$=8 in our implementation). The class probability can be decoupled into a target binary probability $P_b$ (whether belonging to the target class, TAKD) and a non-target multi-class probability $P_m$ (which non-target category it belongs to, NAKD). We employ KL-divergence to measure the teacher-student similarity, facilitating knowledge transfer from TSkel-Mamba with CP to TSkel-Mamba with GAP. As shown in \cref{tab1}, CPKD not only achieves further performance gains but also maintains the original parameter count.

\textbf{Discussion of CPKD.} CPKD is a practical optimization strategy for skeleton-based action recognition, featuring a two-stage processing pipeline. We openly acknowledge that CPKD introduces additional training overhead. However,  we believe that 1) \textsl{Inference Efficiency is the Core Focus.} Our emphasis on efficiency pertains primarily to inference time, which is critical for real-world applications such as robotics, VR, and edge computing, where models are deployed in resource-constrained environments. While CPKD involves a two-stage training process, it introduces no additional overhead during inference. Therefore, the overall inference cost remains unchanged, even when CPKD is used during training. We argue that a slight increase in training complexity is a worthwhile trade-off when it leads to improved performance without altering the inference-time, model size, FLOPs, or latency; 2) \textsl{Minimal Training Overhead and Practical Benefits.} CPKD is a parameter-free logit distillation method, which makes it lighter than conventional knowledge distillation strategies. It helps accelerate convergence of the student model and enhances generalization, all while adding minimal extra computation during training. Although CPKD incurs some extra training overhead, its inference performance and accuracy gains make it a worthwhile optimization strategy for action recognition.

\section{Experiments}
\label{Sec4}

This section compares the advancement of our TSkel-Mamba with state-of-the-arts and conducts ablation studies to  verify the effectiveness of the proposed approaches.

\subsection{Datasets}

We evaluate our proposed method on four widely used  action  datasets: NTU-RGB+D 60 (NTU60), NTU-RGB+D (NTU120),  Northwestern-UCLA (NW-UCLA) and UAV-Human.

\textbf{1)} \textbf{NTU-RGB+D 60}~\cite{NTU60}  is an authoritative human action dataset containing 56,880 samples, including 25 major human joints. Two evaluation benchmarks: (a) \textit{Cross-View} (\textit{X-View}). Different camera views ($-45^{\circ}$, $0^{\circ}$, $+45^{\circ}$)  are positioned on the same horizontal line. A total of 37,920 samples are collected from cameras 2 and 3 for training, while 18,960 samples are obtained from camera 1 for testing. (b) \textit{Cross-Subject} (\textit{X-Sub}).  40 subjects performed a total of 60 different types of actions (40 daily, 11 interactive, and 9 health-related), yielding 40,320 training samples and 16,560 test samples.

\textbf{2)} \textbf{NTU-RGB+D 120}~\cite{NTU120} extends NTU60, totaling 114,480 samples from 106 subjects across 120 classes. Two  benchmarks: (a) \textit{Cross-Subject} (\textit{X-Sub}), where actions from 53 subjects are used for training and the rest for testing;   (b) \textit{Cross-Setup} (\textit{X-Set}), where samples with even setup IDs are used for training and those with odd IDs for testing.

\textbf{3)} \textbf{Northwestern-UCLA }~\cite{NUCLA}  is another common action dataset, including 20  human joints. The evaluation benchmark follows the same criteria as NTU60, using samples from two cameras for training and the rest for testing.

\textbf{4) UAV-Human}~\cite{UAV}   is another large-scale action dataset, comprising 22,476 video clips across 155 distinct categories. Captured by UAVs traversing urban and rural environments under varying illumination conditions, the dataset is partitioned into 89 subjects for training and 30 for testing, maintaining a challenging benchmark.

\subsection{Implementation Details}
\label{Sec:42}
Our method is implemented  by \textit{Python} and \textit{Pytorch}, and trained on a single \textit{RTX 4090} GPU. The code and data preprocessing strategy  are based on \cite{Mamba,CTR-GCN,HyperFormer}. Each action sample is resized to 64 frames. The training script is based on \cite{CTR-GCN,HyperFormer}.  We employed the \textit{stochastic gradient descent} (SGD) optimizer with a weight decay of 0.0004 to avoid overfitting.  The initial learning rate is set at 0.025, with a decay rate of 0.1 during the 110$th$ to 120$th$ epochs. The batch size is opted to 64 in both NTU60 and NTU120, and 16 in NW-UCLA. In addition, our TSkel-Mamba utilizes a 7-layer \textit{spatial-temporal} architecture with 216 channels.  The number of heads in spatial Transformer is set to 9. 



\begin{table*}[!]
	\caption{
		Comparison of the accuracy with state-of-the-arts on NTU60, NTU120 and NW-UCLA datasets. $\mathbb{S}_1$ and $\mathbb{S}_2$ mean the \textit{joint}-stream  and the \textit{two}-stream. $\mathbb{S}_4$ means \textit{four}-stream. \textbf{Bold} highlights the best performance. \textit{S} means spatial. \textit{ST} mean spatial and temporal. CPKD is an optimization strategy using covariance pooling and logit-based knowledge distillation, described in \cref{Sec:42}.}
	\resizebox{\linewidth}{!}{
 
	\begin{tabular}{ccccccc|ccc|ccc|ccc|c}
 
        \toprule
             & & &   &\multicolumn{6}{c}{\textbf{NTU-RGB+D 60}}  & \multicolumn{6}{c}{\textbf{NTU-RGB+D 120}}  & \textbf{NW-UCLA} \\ 
		\multirow{2}{*}{\textbf{Types}}& \multirow{2}{*}{\textbf{Methods}} &\multirow{2}{*}{\textbf{Years}} & \multirow{2}{*}{\textbf{Params}}  & \multicolumn{3}{c}{\textsl{X-Sub(\%)}}&  \multicolumn{3}{c}{\textsl{X-View(\%)}}& \multicolumn{3}{c}{\textsl{X-Set(\%)}}                               & \multicolumn{3}{c}{\textsl{X-Sub(\%)}}  &   \textsl{Top-1(\%)}  \\ 
  
	& & &  
        & $\mathbb{S}_1$    & $\mathbb{S}_2$
        & $\mathbb{S}_4$    & $\mathbb{S}_1$     
        & $\mathbb{S}_2$    & $\mathbb{S}_4$  
        & $\mathbb{S}_1$    & $\mathbb{S}_2$    
        & $\mathbb{S}_4$    & $\mathbb{S}_1$     
        & $\mathbb{S}_2$    & $\mathbb{S}_4$

        \\\hline\hline
        
         \multirow{1}{*}{CNN}& Ta-CNN~\cite{Ta-CNN}& AAAI`22  & $-$ 
         &88.8 &$-$ &90.4 
         &93.6 &$-$ &94.8   
         &84.0 &$-$ &86.8   
         &82.4 &$-$ &85.7 
         &96.1 \\ \hline

         \multirow{10}{*}{GCN}& SGN~\cite{SGN}& CVPR`20  & $-$ 
         &$-$ &89.0 &$-$ 
         &$-$ &94.5 &$-$  
         &$-$ &81.5 &$-$  
         &$-$ &79.2 &$-$
         &$-$ \\ 

         & MS-G3D~\cite{MS-G3D} & CVPR`20  & 2.8M 
         &89.4 &91.5 &$-$ 
         &94.9 &96.2 &$-$
         &84.4 &88.4 &$-$ 
         &83.3 &86.9 &$-$
         &$-$ \\  
         
         &CTR-GCN~\cite{CTR-GCN} & ICCV`21  &1.4M
         &89.9 &92.2 &92.4 
         &$-$ &$-$ &96.8   
         &86.4 &90.1 &90.6   
         &84.9 &88.7 &88.9
         &96.5 \\ 

         &MST-GCN~\cite{MST-GCN} & AAAI`21  &12.0M
         &89.0 &91.1 &91.5 
         &95.1 &96.4 &96.6  
         &84.5 &88.3 &88.8   
         &82.8 &87.0 &87.5
         &$-$ \\ 
         
         &ST-GCN++~\cite{STGCNplus} & MM`22  &1.4M
         &89.3 &91.4 &92.1 
         &95.6 &96.7 &97.0  
         &85.6 &87.5 &89.8   
         &83.2 &87.0 &87.5
         &$-$ \\ 
         
         &Info-GCN~\cite{InfoGCN} & CVPR`22  &1.6M
         &89.8 &91.6 &92.3
         &95.2 &96.5 &96.7  
         &86.3 &89.7 &90.7   
         &85.1 &88.5 &89.2
         &96.6 \\ 
         
         &HD-GCN~\cite{HD-GCN} & ICCV`23  &1.7M
         &90.6 &92.4 &93.0
         &95.7 &96.6 &97.0  
         &87.3 &90.6 &91.2   
         &85.7 &89.1 &89.7
         &96.9 \\

         &FR-Head~\cite{FR-Head} & CVPR`23  &1.7M
         &90.3 &92.3 &92.8
         &95.3 &96.4 &96.8  
         &87.3 &$-$ &90.9   
         &85.5 &$-$ &89.5
         &96.8 \\
    
         &Block-GCN~\cite{BlockGCN} & CVPR`24  &1.4M
         &90.9 &$-$ &93.1
         &95.4 &$-$ &97.0 
         &88.2 &$-$ &91.5   
         &86.9 &$-$ &90.3
         &96.9 \\ \hline

         \multirow{1}{*}{\textit{S}-Transformer}& Hyperformer~\cite{HyperFormer}& $-$  & 2.7M 
         &90.7 &$-$ &92,9 
         &95.1 &$-$ &96.5  
         &88.0 &$-$ &91.3 
         &86.6 &$-$ &89.9
         &96.7 \\ \hline

         \multirow{4}{*}{\textit{ST}-Transformer}& DSTA-Net~\cite{DSTA-Net}& ACCV`20  & 4.1M
         &$-$ &$-$  &91.5 
         &$-$ &$-$  &96.4  
         &$-$ &$-$  &89.0  
         &$-$ &$-$  &86.6
         &$-$ \\ 

         &ST-TR~\cite{ST-TR} &ICPR`21  & 12.1M 
         &88.7 &89.9 &$-$ 
         &95.6 &96.1 &$-$  
         &$-$ &84.1 &$-$
         &$-$ &81.9 &$-$
         &$-$ \\ 

        &STST~\cite{STST} & MM`21  & $-$ 
         &$-$ &$-$ &91.9 
         &$-$ &$-$ &96.8 
         &$-$ &$-$ &$-$  
         &$-$ &$-$ &$-$
         &$-$ \\ 



        & FG-STFormer~\cite{FG-STFormer} & ACCV`22  &$-$
         &$-$ &$-$ &92.6 
         &$-$ &$-$ &96.7  
         &$-$ &$-$ &90.6  
         &$-$ &$-$ &89.0
         &97.0 \\ \hline

         \multirow{1}{*}{Text Descriptions}& GAP~\cite{GAP} & ICCV`23  &2.1M
         &90.2 &$-$ &92.9
         &95.6 &$-$ &97.0  
         &87.0 &$-$ &91.1   
         &85.5 &$-$ &89.9
         &\textbf{97.2} \\ \hline

        \multirow{5}{*}{\makecell[c]{Temporal \\ Modeling}} & DPRL-GCN~\cite{DPRL-GCN}& CVPR`18  & $-$
         &83.5 &$-$  &$-$ 
         &89.8 &$-$  &$-$ 
         &$-$ &$-$  &$-$  
         &$-$ &$-$  &$-$
         &$-$ \\

        & AGC-LSTM~\cite{AGC-LSTM}& CVPR`19  & $-$ 
         &87.5 &89.2 &$-$ 
         &93.5 &95.0 &$-$   
         &$-$ &$-$ &$-$   
         &$-$&$-$ &$-$ 
         &93.3 \\

        & TCA-GCN~\cite{TCA-GCN} & $-$  & 2.6M 
         &$-$ &$-$ &92.9 
         &$-$ &$-$ &97.0  
         &$-$ &$-$ &90.8 
         &85.3 &$-$ &89.4
         &$-$ \\ 

        & Koopman~\cite{Koopman}& CVPR`23  & 5.3M 
         &90.2 &$-$ &92.9 
         &95.2 &$-$ &96.8  
         &87.5 &$-$ &91.3 
         &85.7 &$-$ &90.0
         &97.0 \\ 
\rowcolor{gray!20}
        & TSkel-Mamba& Ours  & 2.4M 
         &91.4 &92.9 &93.1 
         &95.8 &96.8 &97.2  
         &88.7 &90.9 &91.6  
         &87.4 &89.7 &90.4 
         &97.0 \\ 
\rowcolor{gray!20}
        & \textbf{TSkel-Mamba (w/ CPKD)}& \textbf{Ours}  & 2.4M 
         &\textbf{91.6} &\textbf{93.0} &\textbf{93.2}
         &\textbf{96.2} &\textbf{97.1} &\textbf{97.4}  
         &\textbf{88.9 } &\textbf{91.0} &\textbf{91.7 } 
         &\textbf{87.9} &\textbf{90.0} &\textbf{90.6} 
         &\textbf{97.2 }\\ 
         
        \bottomrule 
        
    \end{tabular}}
    \label{tab1}
    \vspace{-13pt}
\end{table*}

\begin{table}[!]
\centering
\caption{Comparison of the accuracy with state-of-the-arts on UAV-Human dataset}
\label{tab:UAV}
\begin{tabular}{ccc}
\toprule
\textbf{Methods} & \textbf{Years} & \textbf{CSv1 (\%)} \\
\midrule
DGNN\cite{DGNN} & CVPR'19 & 29.9 \\
ST-GCN\cite{ST-GCN} & AAAI'18 & 30.3 \\
2s-AGCN\cite{2s-AGCN} & CVPR'19 & 34.8 \\
Shift-GCN\cite{ShiftGCN} & CVPR'20 & 38.0 \\
CTR-GCN\cite{CTR-GCN} & ICCV'21 & 43.4 \\
MKE-GCN\cite{MKEGCN} & ICME'22 & 44.6 \\
ACFL-CTR\cite{ACFL} & MM'22 & 44.2 \\
Koopman\cite{Koopman} & CVPR'23 & 44.2 \\
TD-GCN\cite{TDGCN} & TMM'24 & 45.4 \\\hline\rowcolor{gray!20}
\textbf{TSkel-Mamba} & \textbf{Ours} & \textbf{47.2} \\
\bottomrule
\end{tabular}
\end{table}

\subsection{Multi-stream Strategy }
The use of multi-stream input modalities is a widely adopted and standardized practice in skeleton-based action recognition\cite{2s-AGCN, CTR-GCN, HyperFormer, BlockGCN}. Specifically:

1) The joint-stream uses raw 3D joint coordinates.
2) The bone-stream encodes relative displacements between adjacent joints.
3) The joint-motion stream captures temporal changes in joint positions.
4) The bone-motion stream models temporal bone displacements.

In most mainstream settings, joint and bone streams are fused via late softmax score summation to form the two-stream configuration. Similarly, the fusion of all four streams constitutes the four-stream setting. These configurations serve distinct purposes: 

1) The joint-stream setting $\mathbb{S}_1$ is essential for evaluating the core temporal modeling capacity of the backbone using raw input. 

2) The two-stream  setting $\mathbb{S}_2$ adds relational spatial context (bone) for performance enhancement without significant computational burden. 

3) The four-stream setting $\mathbb{S}_4$ combines all the four streams and is thus powerful.

\subsection{Comparison with the State-of-the-arts}
\cref{tab1} compares the recognition accuracy of our TSkel-Mamba with  state-of-the-art approaches on the NTU60, NTU120 and  NW-UCLA dataset.  We have also integrated  multi-stream strategies~\cite{2s-AGCN,MS-G3D}, including \textit{joint}, \textit{bone}, \textit{motion}  and \textit{motion-bone} streams.  

\textbf{Spatial-Dominated Approaches.} Our approach outperforms Ta-CNN~\cite{Ta-CNN} by 4.9\% and 5.5\% on the NTU120 \textit{X-Set} and \textit{X-Sub} benchmarks.  Compared to the GCN-based Block-GCN~\cite{BlockGCN}, our method achieves superior recognition performance,  +0.7\% and +1.0\% on the NTU60 and NTU120 \textit{XSub} under the \textit{joint}-stream. Moreover, our method significantly surpasses the spatial Transformer based Hyperformer~\cite{HyperFormer} by 0.9\% and 1.3\% in accuracy on the NTU120 \textit{XSet} and \textit{XSub}. While these methods incorporate temporal convolutions, their performance gains are primarily driven by enhancements in spatial modeling rather than effective temporal representation.

\textbf{Temporal Modeling Approaches.} There are only a few temporal modeling approaches due to insufficient exploration. Our method outperforms TCA-GCN~\cite{TCA-GCN} by 0.9\% and 1.2\% on  the NTU120 \textit{XSet} and \textit{XSub} benchmarks under the \textit{four}-stream.  Compared to the latest temporal Koopman pooling~\cite{Koopman},  we also  achieved significant improvements of  1.4\% and 1.0\%  on the NTU60 \textit{XSub} and \textit{XView}, and 1.4\% and 2.2\%  on  NTU120 \textit{XSet} and \textit{XSub} under the \textit{joint}-stream, requiring fewer parameters.

\textbf{TDM vs. Temporal Transformer}.  We have   collected several existing temporal Transformer based recognition methods~\cite{DSTA-Net,ST-TR,STST,FG-STFormer}.  Our TSkel-Mamba significantly outperforms the representative ST-TR~\cite{ST-TR} by 8.1\% on the NTU120 \textit{XSub} benchmarks under the \textit{two}-stream,  with approximately 5$\times$ fewer parameters. Compared with recent FG-STFormer~\cite{FG-STFormer}, our method also achieves better recognition performance (+1.1\% and +1.6\% on the NTU120 \textit{XSet} and \textit{XSub} under the \textit{four}-stream.)

\textbf{Challenging Datasets. }Observation reveals that the challenges inherent in the dataset result in less pronounced performance gains. We would like to emphasize that NTU-RGB D is a highly saturated and competitive benchmark, with over 100,000 labeled skeleton sequences. In this context, even 0.1\% gain equates to roughly 100 additional correctly classified samples, making marginal improvements statistically and practically meaningful—a pattern widely accepted in the field.  To further validate our method's effectiveness, we conducted additional comparisons on UAV-Human - another large-scale action dataset, comprising 22,476 video clips across 155 distinct categories, maintaining a challenging benchmark. With more significant margin, our TSkel-Mamba achieves a significant 3\% performance gain over latest Koopman\cite{Koopman} as shown in \cref{tab:UAV}.

\subsection{Ablation Studies}
In this section, we conduct ablation studies to validate the effectiveness of the proposed method and its internal modules.

\textbf{Baseline.} To ensure a fair comparison, we implemented a  \textit{Baseline} model that retains the spatial-temporal architecture. For the spatial component, we employed the same spatial Transformer configuration used in our proposed TSkel-Mamba.  For temporal modeling, we adopted the widely used Temporal Convolutional Network (TCN), following the implementations in prior works \cite{CTR-GCN,FR-Head,BlockGCN}. The consistent performance improvements over this baseline underscore the effectiveness of our proposed TDM block, demonstrating its superior ability to capture complex temporal dependencies compared to traditional TCNs.

\begin{table}[!]
    \centering
    \caption{ Validation of the effectiveness of  TDM and Internal Components on NTU120 X-Sub.  $w/o$  MS denotes the replacement of multi-scale (MS) $S_{K}=\{1,3,5\}$ Cycle  by single scale with $S_{K}=\{3\}$. $\mathbb{S}_1$ means the \textit{joint}-stream.}
    \begin{tabular}{lc}
    \toprule
    \multicolumn{1}{c}{\textbf{Method}}   & \textbf{Acc} (\%) $\mathbb{S}_1$\\ \hline
    \textit{Baseline}                     &84.2     \\ \hline
    \textit{T-Scan} Mamba                &85.8$^{+1.6}$ \\
    \textit{T-Scan} Mamba + MTI $w/o$  MS                    &86.4$^{+2.2}$     \\
    \textit{T-Scan} Mamba + MTI                              &86.7$^{+2.5}$     \\ \rowcolor{gray!20}
    \textbf{TDM}                              &\textbf{87.4}\textcolor{red}{$^{+3.2}$}  \\
        \bottomrule  
    \end{tabular}
    \label{tab2}
\end{table}

\begin{table}[!]
        \centering
         \caption{ Comparison of different scanning strategies for Mamba on NTU120 XSub. $\mathbb{S}_1$ denotes the joint stream. $w/ X$-Mamba denotes that each layer of the \textit{baseline} integrated a  Mamba block with $X$ scanning strategy. }
        \begin{tabular}{lc}
        \toprule
        \multicolumn{1}{c}{\textbf{Method}} & \textbf{Acc} (\%) $\mathbb{S}_1$\\ \hline
        \textit{Baseline}                 &84.2  \\  \hline
            $w/$ \textit{TS-Scan}  Mamba  &85.3  \\
            $w/$ \textit{ST-Scan}  Mamba  &85.5  \\ 
            $w/$ \textit{S-Scan}   Mamba  &84.6  \\ \rowcolor{gray!20}
            \textbf{\textit{T-Scan} Mamba}    &\textbf{85.8$^{+1.6}$} \\ \bottomrule  
        \end{tabular}
        \label{tab3}
\end{table}

\textbf{Effectiveness of Internal Components in TDM.} As shown in~\cref{tab2}, to verify the benefit of using mamba for temporal modeling, we first attempt to replace the TCN in the baseline with the  \textit{Temporal-prioritized} scanning mamba while retaining the spatial module, thereby constructing a simple mamba-based model, named \textit{T-Scan}  Mamba, with improvements of +1.6\% in accuracy  than \textit{Baseline}  on the NTU120-XSub.  Subsequently, we validated the effectiveness of both single-scale ($S_{K}=\{3\}$) and multi-scale ($S_{K}=\{1,3,5\}$)  MTI modules, achieving performance improvements of 2.2\% and 2.5\%, respectively.  Finally, we validated the effectiveness of our proposed TDM framework. Encouragingly, it demonstrated a significant improvement of 3.2\% over the \textit{Baseline}, showing the exceptional capability of TDM in modeling the temporal dynamics of the skeleton action.

\textbf{Impact of Different Scanning Strategies.}  \cref{tab3} compares the effectiveness of mamba with different scanning strategies.   \textit{Temporal-only} scanning helps the \textit{Baseline} achieve a significant improvement of  1.6\% on NTU120 X-Sub, outperforming \textit{Temporal-prioritized} \textit{TS-Scan} by 0.5\% on NTU120 X-Sub,  \textit{Spatial-prioritized}  \textit{ST-Scan} by 0.3\% on NTU120 X-Sub and \textit{Spatial-only} \textit{S-Scan} by  1.2\% on NTU120 X-Sub.   It is more robust for generating scanning  sequences that maintain natural temporal correlations, thus achieving superior performance gains.

\begin{table}[!]
    \centering
    \caption{ Different Temporal-Channel Interaction in  MTI  on NTU120 XSub.  C-Agg denotes feature aggregation along channel dimension. T-Agg  denotes feature aggregation along temporal dimension with different kernel sizes. MS denotes multi-scale. }
    \begin{tabular}{lcc}
        \toprule
        \multicolumn{1}{c}{\textbf{Method}}   & \textbf{Acc} (\%) $\mathbb{S}_1$\\ \hline
        \textit{Baseline}                    &84.2    \\ \hline
            TDM-MTI $w/o$ C-Agg                      &86.2    \\
            TDM-MTI $w/o$ T-Agg, MS                   &86.6     \\
            TDM-MTI $w/o$ MS                           &87.1     \\ \rowcolor{gray!20}
            \textbf{TDM}                            &\textbf{87.4}     \\
            \bottomrule  
        \end{tabular}
    \label{tab4}
\end{table}

\begin{table}[!]
    \centering
    \caption{ Compatibility comparisons with  state-of-the-art spatial-dominant methods on NTU120 XSub.  * denotes the accuracy  reproduced with official code. }
    \begin{tabular}{lcc}
    \toprule
    \multicolumn{1}{c}{\textbf{Method}} &\textbf{Params}  & \textbf{Acc} (\%) $\mathbb{S}_1$\\ \hline
    CTR-GCN~\cite{CTR-GCN} &1.4M  & 84.9 \\
    \textbf{CTR-GCN $w/$ TDM} &\textbf{1.7M}  &\textbf{85.4$^{+0.5}$} \\\hline\hline
    Block-GCN~\cite{BlockGCN}&1.4M   &86.5* \\
    \textbf{Block-GCN $w/$ TDM} &\textbf{1.8M}  &\textbf{87.3$^{+0.8}$}  \\\hline\hline
    Hyperformer~\cite{HyperFormer} & 2.7M  &86.4* \\
    \textbf{Hyperformer $w/$ TDM} & \textbf{3.0M}  &\textbf{87.3$^{+0.9}$} \\ \bottomrule    
    \end{tabular}
    \label{tab5}
\end{table}

\textbf{Temporal-Channel Interaction in  MTI.}  Cycle FC layers with different temporal kernel sizes  enable the MIT to capture channel interactions at multiple temporal scales simultaneously.   \cref{tab4} compared different interactions strategies. Performing only multi-scale (MS, $S_{K}=\{1,3,5\}$)  temporal aggregation ($w/o$ C-Agg) and only channel aggregation ($w/o$ T-Agg, MS)  resulted in performance improvements of 2.0\% and 2.4\%  over \textit{Baseline}, respectively. In our TDM, the integration of temporal-channel aggregation led to a more significant performance boost of 3.2\% over \textit{Baseline}.  

\textbf{Is TDM a worthwhile temporal plugin?} While we have proposed an effective solution for skeleton-based action recognition-TSkel-Mamba, the validation of Temporal Dynamics Modeling (TDM) as a worthwhile temporal plugin significantly elevates the broader impact of our work. To further assess the generalizability of the TDM block, \cref{tab5} presents its integration with three representative spatial-dominant methods.   TDMs are directly integrated into the early layers of each model.} 

Specifically, given TDM`s  excellent capability in modeling complex temporal dynamics, a key principle of the plugin is to maximize its exposure to long-range sequences.  Most existing spatial-dominant methods adopt multi-layer spatiotemporal architectures with temporal downsampling operations ($stride$=2). Therefore, TDMs should be directly integrated  as  a temporal expert into the layers prior to the temporal downsampling ($stride$=2),  which can capture complete temporal features while maintaining the native architecture. Within specific architectural configurations, multi-layer TDM-TCN hybrid backbone can achieve significant performance improvements while maintaining competitive parameter counts.

Following this principle, \cref{tab5} demonstrates that our TDM plugin significantly improves recognition performance and shows excellent compatibility with three state-of-the-art spatial-dominant methods. We evaluate on three representative methods: 1) CTR-GCN~\cite{CTR-GCN} as the classical GCN-based approach, 2) Block-GCN~\cite{BlockGCN} as the state-of-the-art GCN-based method, and 3) Hyperformer~\cite{HyperFormer} as a leading Transformer-based method. All implementations replace the original TCN modules with our proposed TDM at layers 1-3 (prior to the first temporal downsampling).  This TDM-TCN hybrid architecture achieves optimal accuracy-parameter trade-offs, resulting in consistent performance improvements ( +0.5\% for CTR-GCN~\cite{CTR-GCN},  +0.8\% for Block-GCN~\cite{BlockGCN},and  +0.9\% for Hyperformer~\cite{HyperFormer}. )   While the addition of TDM introduces a modest increase in parameter count, the observed gains in accuracy demonstrate that it is a valuable and effective temporal enhancement module.

\textbf{Effect of Different Scale Settings on MTI. }\cref{tab2} investigated the impact of the MTI module at different scales on providing robust and comprehensive representations for mamba.  We evaluated four combinations: $S_{K}=\{3\}$, $S_{K}=\{1,3\}$, $S_{K}=\{1,3,5\}$ and $S_{K}=\{1,3,5,7\}$. Under the single-scale configuration $S_{K}=\{3\}$, MTI module achieved a 2.2\% performance improvement over \textit{Baseline},  Under the multi-scale configuration $S_{K}=\{1,3,5\}$, the optimal performance was achieved, with a 2.5\% improvement over the \textit{Baseline}.  Adding the $K=7$ scale caused a slight performance drop, likely due to its larger receptive field leading to overly coarse feature extraction.  Therefore, the MTI was applied with the optimal scale configuration $S_{K}=\{1,3,5\}$.

\begin{figure}[!]
	\centering 
	\includegraphics[width=0.49\textwidth]{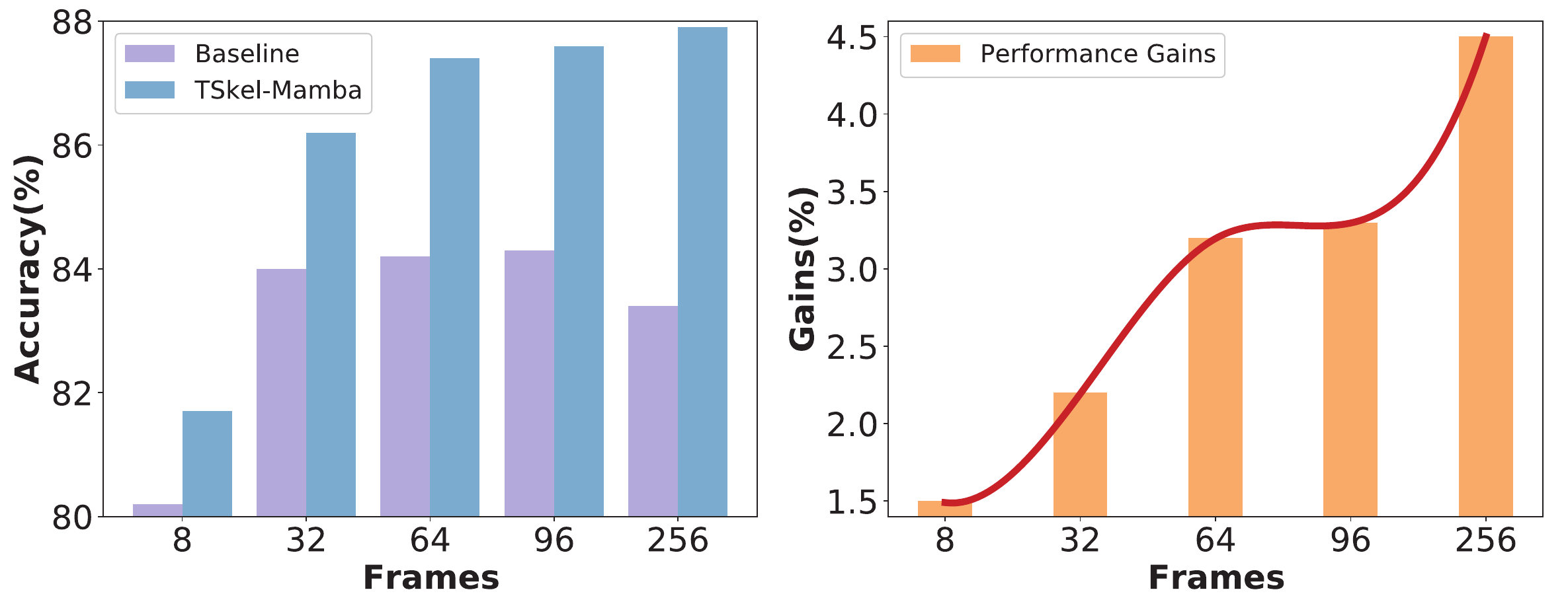}
	\caption{Comparison of  accuracy with different numbers of frames.  “Gains” represents the performance gains from TDM .}
	\label{fig:gains}
\end{figure}

\begin{table}[!]
\centering
\caption{Performance Comparison of the MIT Module at Different Scales on NTU120 XSub. \  $S_{K}$ denotes a  set of  kernel sizes at different scales. }
\begin{tabular}{lcc}
\toprule
\multicolumn{1}{c}{\textbf{Method}} & \textbf{Acc} (\%) $\mathbb{S}_1$\\ \hline
\textit{Baseline}               & 84.2    \\  \hline
    \textit{T-Scan} Mamba-MTI $w/$ $S_{K}=\{3\}$      & 86.4     \\
    \textit{T-Scan} Mamba-MTI $w/$ $S_{K}=\{1,3\}$   & 86.5     \\ 
    \textit{T-Scan} Mamba-MTI $w/$ $S_{K}=\{1,3,5,7\}$     &86.4   \\ \rowcolor{gray!20}
    \textbf{\textit{T-Scan} Mamba-MTI $w/$ $S_{K}=\{1,3,5\}$}   & \textbf{86.7}$^{+2.5}$     \\ \bottomrule  
\end{tabular}

\label{tab6}

\end{table}

\begin{table}[!]
\centering
\caption{Comparison of accuracy and efficiency gains of CPKD.}
\label{tab:tskel-comparison}
\begin{tabular}{cccc}
\toprule
\textbf{Method} & \textbf{Acc (\%)} & \textbf{Param} & \textbf{FLOPs} \\
\midrule
TSkel-Mamba w/ CP & 87.8 & 3.9M & 8.3G \\
light-TSkel-Mamba & 86.5 & 0.9M & 3.5G \\\rowcolor{gray!20}
light-TSkel-Mamba+CPKD & 87.7 & 0.9M & 3.5G \\
\bottomrule
\end{tabular}
\end{table}

\textbf{Potential of CPKD.} CPKD can improve model recognition performance without increasing inference overhead (\cref{tab1}). We further explore its potential for model lightweighting. To demonstrate its practicality, we performed an additional experiment using a compressed variant of our model—light-TSkel-Mamba (3-layer version). When trained with CPKD, this lightweight student model achieves a remarkable improvement in accuracy while retaining high computational efficiency as shown in \cref{tab:tskel-comparison}. This highlights how CPKD can be effectively used to train smaller and more efficient variants of TSkel-Mamba, further reinforcing its value in low-resource deployment scenarios.

\textbf{Long-Term Action Sequences.} Our proposed TSkel-Mamba focuses on modeling the temporal dynamics of skeleton-based actions. Therefore, it is essential to investigate the impact of varying numbers of skeleton frames on recognition performance. Note that the original skeleton sequences contain 300 frames, where most prior work~\cite{CTR-GCN,Koopman,GAP} adopts random cropping with bilinear interpolation to downsample to 64 frames. Guided by the same cropping operation,  \cref{fig:gains} tests the sensitivity of the proposed TSkel-Mamba with respect to different sequence lengths, sampling actions such as frames $[8, 32, 64, 96,256]$ by controlling the window size.

We observed that the performance of \textit{Baseline}  initially increases with more frames, then decreases, suggesting increasingly complex temporal correlations.  Interestingly, the performance of our TSkel-Mamba improves with increasing frame numbers. However, TSkel-Mamba exhibits diminishing returns as the frame count reaches 256. Significant performance degradation of \textit{ Baseline} suggests that excessive frames introduce redundant temporal information, which is not conducive to capturing temporal dynamics.  To quantify this,  \cref{fig:gains} (right) compares the relative performance gains against the \textit{ Baseline}. The encouraging gains validates that TDM can robustly capture temporal dynamics from more complex and longer sequences, showing potential for long-term action recognition.

\textbf{Comparison of accuracy under each action label:} \cref{tab333} compares the recognition accuracy based on each action label, providing an in-depth analysis of our proposed TDM`s effectiveness. Specifically, we compared the recognition accuracy between the $Baseline$ and $Baseline$ with TDM block based on action labels. Additionally, we calculated TDM`s performance gain per action label (with \textcolor{red}{red} and \textcolor{blue}{blue} indicating positive and negative impacts, respectively). The experiments were conducted on the NTU120-XSub, which includes 120 action categories in total. TDM demonstrated negative effects on only 10 action label (average gain: -0.94\%), \textbf{positive} improvements on \textbf{105} labels (average gain: +\textbf{3.76}\%), and no impact on 5 labels. further validating the effectiveness of TDM.

\textbf{TDM's performance on hard actions:} Firstly, we define $hard$ action based on the $Baseline$`s recognition accuarcy. Specifically, the bottom ten least-accurate action labels ( \textcolor{violet}{purple} entries  in \cref{tab3} ) are regarded as $hard$ actions. Notably, TDM shows surprisingly positive effects on all $hard$ labels, achieving a maximum improvement of \textbf{16.18}\% for label $A12$ and an average gain of \textbf{9.38}\% compared to the $Baseline$, as shown in \cref{tab:R55}. TDM enables effective recognition of $hard$ actions, leading to comprehensive accuracy improvements.

\begin{table*}[!]
\centering
\caption{Comparison of TDM's performance gains on hard actions}
\label{tab:R55}
\begin{tabular}{cccc}
\toprule
& \multicolumn{1}{c}{\textbf{Average Positive Gain/Number}} & \multicolumn{1}{c}{\textbf{Maximum Gain of Hard Action}} & \multicolumn{1}{c}{\textbf{Average Gain of Hard Action}} \\
\midrule
\textbf{Improvement} & +3.76\%/105 & \textbf{16.18}\% & \textbf{9.38}\% \\
\bottomrule
\end{tabular}
\end{table*}

\begin{table}[!]
\centering
\caption{Efficiency comparison of different methods}
\label{tab:FLOP}
\begin{tabular}{cccc}
\toprule
\textbf{Methods} & \textbf{Params (M)} & \textbf{FLOPs (G)} & \textbf{Acc (\%)} \\
\midrule
ST-TR\cite{ST-TR} & 12.1 & 259.4 & 82.7 \\
DSTA-Net\cite{DSTA-Net} & 3.4 & 16.2 & 84.0 \\
Hyperformer\cite{HyperFormer} & 2.7 & 9.6 & 86.6 \\
Koopman\cite{Koopman} & 5.3 & 8.8 & 85.7 \\\rowcolor{gray!20}
\textbf{TSkel-Mamba (Ours)} & \textbf{2.4} & \textbf{8.2 }& \textbf{87.9} \\
\bottomrule
\end{tabular}
\end{table}

\begin{table*}[!]
\centering
\caption{Action label descriptions and comparison of recognition accuracy under each action label on NTU120 XSub. \textcolor{red}{red} and \textcolor{blue}{blue} indicating positive and negative impacts and purple, \textcolor{violet}{purple} entries highlights the $hard$ action (the bottom ten least-accurate action labels).   }
\resizebox{\linewidth}{!}{
\renewcommand{\arraystretch}{0.75}
\begin{tabular}{@{}llll@{}}
\toprule
\multicolumn{4}{c}{\textbf{Action  Label Description}}\\\hline
{\color[HTML]{333333} A1.   drink water.}                 & {\color[HTML]{333333} A31. pointing to something with   finger.}           & {\color[HTML]{333333} A61. put on headphone.}                 & {\color[HTML]{333333} A91. open a box.}                                            \\
{\color[HTML]{333333} A2. eat   meal/snack.}              & {\color[HTML]{333333} A32. taking a selfie.}                               & {\color[HTML]{333333} A62. take off headphone.}               & {\color[HTML]{333333} A92. move heavy objects.}                                    \\
{\color[HTML]{333333} A3.   brushing teeth.}              & {\color[HTML]{333333} A33. check time (from watch).}                       & {\color[HTML]{333333} A63. shoot at the basket.}              & {\color[HTML]{333333} A93. shake fist.}                                            \\
{\color[HTML]{333333} A4.   brushing hair.}               & {\color[HTML]{333333} A34. rub two hands together.}                        & {\color[HTML]{333333} A64. bounce ball.}                      & {\color[HTML]{333333} A94. throw up cap/hat.}                                      \\
{\color[HTML]{333333} A5.   drop.}                        & {\color[HTML]{333333} A35. nod head/bow.}                                  & {\color[HTML]{333333} A65. tennis bat swing.}                 & {\color[HTML]{333333} A95. hands up (both hands).}                                 \\
{\color[HTML]{333333} A6.   pickup.}                      & {\color[HTML]{333333} A36. shake head.}                                    & {\color[HTML]{333333} A66. juggling table tennis   balls.}    & {\color[HTML]{333333} A96. cross arms.}                                            \\
{\color[HTML]{333333} A7.   throw.}                       & {\color[HTML]{333333} A37. wipe face.}                                     & {\color[HTML]{333333} A67. hush (quite).}                     & {\color[HTML]{333333} A97. arm circles.}                                           \\
{\color[HTML]{333333} A8.   sitting down.}                & {\color[HTML]{333333} A38. salute.}                                        & {\color[HTML]{333333} A68. flick hair.}                       & {\color[HTML]{333333} A98. arm swings.}                                            \\
{\color[HTML]{333333} A9.   standing up.}                 & {\color[HTML]{333333} A39. put the palms together.}                        & {\color[HTML]{333333} A69. thumb up.}                         & {\color[HTML]{333333} A99. running on the spot.}                                   \\
{\color[HTML]{333333} A10.   clapping.}                   & {\color[HTML]{333333} A40. cross hands in front (say   stop).}             & {\color[HTML]{333333} A70. thumb down.}                       & {\color[HTML]{333333} A100. butt kicks (kick   backward).}                         \\
{\color[HTML]{333333} A11.   reading.}                    & {\color[HTML]{333333} A41. sneeze/cough.}                                  & {\color[HTML]{333333} A71. make ok sign.}                     & {\color[HTML]{333333} A101. cross toe touch.}                                      \\
{\color[HTML]{333333} A12.   writing.}                    & {\color[HTML]{333333} A42. staggering.}                                    & {\color[HTML]{333333} A72. make victory sign.}                & {\color[HTML]{333333} A102. side kick.}                                            \\
{\color[HTML]{333333} A13.   tear up paper.}              & {\color[HTML]{333333} A43. falling.}                                       & {\color[HTML]{333333} A73. staple book.}                      & {\color[HTML]{333333} A103. yawn.}                                                 \\
{\color[HTML]{333333} A14.   wear jacket.}                & {\color[HTML]{333333} A44. touch head (headache).}                         & {\color[HTML]{333333} A74. counting money.}                   & {\color[HTML]{333333} A104. stretch oneself.}                                      \\
{\color[HTML]{333333} A15.   take off jacket.}            & {\color[HTML]{333333} A45. touch chest.}        & {\color[HTML]{333333} A75. cutting nails.}                    & {\color[HTML]{333333} A105. blow nose.}                                            \\
{\color[HTML]{333333} A16.   wear a shoe.}                & {\color[HTML]{333333} A46. touch back (backache).}                         & {\color[HTML]{333333} A76. cutting paper (using   scissors).} & {\color[HTML]{333333} A106. hit other person with   something.}                    \\
{\color[HTML]{333333} A17.   take off a shoe.}            & {\color[HTML]{333333} A47. touch neck (neckache).}                         & {\color[HTML]{333333} A77. snapping fingers.}                 & {\color[HTML]{333333} A107. wield knife towards other   person.}                   \\
{\color[HTML]{333333} A18.   wear on glasses.}            & {\color[HTML]{333333} A48. nausea or vomiting   condition.}                & {\color[HTML]{333333} A78. open bottle.}                      & {\color[HTML]{333333} A108. knock over other person.}            \\
{\color[HTML]{333333} A19.   take off glasses.}           & {\color[HTML]{333333} A49. use a fan (with hand or  paper).} & {\color[HTML]{333333} A79. sniff (smell).}                    & {\color[HTML]{333333} A109. grab other person’s stuff.}                            \\
{\color[HTML]{333333} A20.   put on a hat/cap.}           & {\color[HTML]{333333} A50. punching/slapping other   person.}              & {\color[HTML]{333333} A80. squat down.}                       & {\color[HTML]{333333} A110. shoot at other person with   a gun.}                   \\
{\color[HTML]{333333} A21.   take off a hat/cap.}         & {\color[HTML]{333333} A51. kicking other person.}                          & {\color[HTML]{333333} A81. toss a coin.}                      & {\color[HTML]{333333} A111. step on foot.}                                         \\
{\color[HTML]{333333} A22.   cheer up.}                   & {\color[HTML]{333333} A52. pushing other person.}                          & {\color[HTML]{333333} A82. fold paper.}                       & {\color[HTML]{333333} A112. high-five.}                                            \\
{\color[HTML]{333333} A23.   hand waving.}                & {\color[HTML]{333333} A53. pat on back of other   person.}                 & {\color[HTML]{333333} A83. ball up paper.}                    & {\color[HTML]{333333} A113. cheers and drink.}                                     \\
{\color[HTML]{333333} A24.   kicking something.}          & {\color[HTML]{333333} A54. point finger at the other   person.}            & {\color[HTML]{333333} A84. play magic cube.}                  & {\color[HTML]{333333} A114. carry something with other   person.}                  \\
{\color[HTML]{333333} A25.   reach into pocket.}          & {\color[HTML]{333333} A55. hugging other person.}                          & {\color[HTML]{333333} A85. apply cream on face.}              & {\color[HTML]{333333} A115. take a photo of other   person.}                       \\
{\color[HTML]{333333} A26.   hopping.} & {\color[HTML]{333333} A56. giving something to other   person.}            & {\color[HTML]{333333} A86. apply cream on hand back.}         & {\color[HTML]{333333} A116. follow other person.}                                  \\
{\color[HTML]{333333} A27.   jump up.}                    & {\color[HTML]{333333} A57. touch other person's   pocket.}                 & {\color[HTML]{333333} A87. put on bag.}                       & {\color[HTML]{333333} A117. whisper in other person’s   ear.}                      \\
{\color[HTML]{333333} A28.   make a phone call.}          & {\color[HTML]{333333} A58. handshaking.}                                   & {\color[HTML]{333333} A88. take off bag.}                     & {\color[HTML]{333333} A118. exchange things with other   person.}                  \\
{\color[HTML]{333333} A29.   playing with phone.}  & {\color[HTML]{333333} A59. walking towards each other.}                    & {\color[HTML]{333333} A89. put something into a bag.}         & {\color[HTML]{333333} A119. support somebody with   hand.}                         \\
{\color[HTML]{333333} A30.   typing on a keyboard.}       & {\color[HTML]{333333} A60. walking apart from each   other.}               & {\color[HTML]{333333} A90. take something out of a   bag.}    & {\color[HTML]{333333} A120. finger-guessing game .}\\\bottomrule     
\end{tabular}}
\resizebox{\linewidth}{!}{
\renewcommand{\arraystretch}{0.75}
\begin{tabular}{@{}lcccc||lcccc@{}}
\toprule
\textbf{Classes}                    & \textbf{Baseline} & \textbf{$w/$ TDM} & \textbf{Gains} & \textbf{Baseline/Gains Rank} & \textbf{Classes}                     & \textbf{Baseline} & \textbf{$w/$ TDM} & \textbf{Gains} & \textbf{Baseline/Gains Rank} \\ \hline
{\color[HTML]{333333} \textbf{A1}}  & 0.803             & 0.876          & \textcolor{red}{$\uparrow$7.30\%}         & 32/15         & {\color[HTML]{333333} \textbf{A61}}  & 0.906             & 0.929          & \textcolor{red}{$\uparrow$2.32\%}         & 67/63         \\
{\color[HTML]{333333} \textbf{A2}}  & 0.720             & 0.745          & \textcolor{red}{$\uparrow$2.55\%}         & 20/62         & {\color[HTML]{333333} \textbf{A62}}  & 0.898             & 0.899          & \textcolor{red}{$\uparrow$0.18\%}         & 60/104        \\
{\color[HTML]{333333} \textbf{A3}}  & 0.821             & 0.835          & \textcolor{red}{$\uparrow$1.47\%}         & 38/77         & {\color[HTML]{333333} \textbf{A63}}  & 0.857             & 0.902          & \textcolor{red}{$\uparrow$4.55\%}         & 46/30         \\
{\color[HTML]{333333} \textbf{A4}}  & 0.875             & 0.923          & \textcolor{red}{$\uparrow$4.76\%}         & 55/28         & {\color[HTML]{333333} \textbf{A64}}  & 0.967             & 0.974          & \textcolor{red}{$\uparrow$0.70\%}         & 104/90        \\
{\color[HTML]{333333} \textbf{A5}}  & 0.865             & 0.862          & \textcolor{blue}{$\downarrow$-0.36\%}       & 50/113        & {\color[HTML]{333333} \textbf{A65}}  & 0.808             & 0.890          & \textcolor{red}{$\uparrow$8.19\%}         & 35/11         \\
{\color[HTML]{333333} \textbf{A6}}  & 0.967             & 0.956          & \textcolor{blue}{$\downarrow$-1.09\%}       & 105/116       & {\color[HTML]{333333} \textbf{A66}}  & 0.955             & 0.974          & \textcolor{red}{$\uparrow$1.92\%}         & 93/68         \\
{\color[HTML]{333333} \textbf{A7}}  & 0.887             & 0.931          & \textcolor{red}{$\uparrow$4.36\%}         & 57/32         & {\color[HTML]{333333} \textbf{A67}}  & 0.770             & 0.773          & \textcolor{red}{$\uparrow$0.35\%}         & 29/98         \\
{\color[HTML]{333333} \textbf{A8}}  & 0.941             & 0.982          & \textcolor{red}{$\uparrow$4.03\%}         & 85/37         & {\color[HTML]{333333} \textbf{A68}}  & 0.783             & 0.871          & \textcolor{red}{$\uparrow$8.87\%}         & 30/10         \\
{\color[HTML]{333333} \textbf{A9}}  & 0.974             & 0.985          & \textcolor{red}{$\uparrow$1.10\%}         & 107/81        & {\color[HTML]{333333} \textbf{A69}}  & 0.692             & 0.736          & \textcolor{red}{$\uparrow$4.35\%}         & 16/33         \\
{\color[HTML]{333333} \textbf{A10}} & 0.806             & 0.864          & \textcolor{red}{$\uparrow$5.86\%}         & 33/25         & {\color[HTML]{333333} \textbf{A70}}  & 0.864             & 0.896          & \textcolor{red}{$\uparrow$3.13\%}         & 49/51         \\
{\color[HTML]{333333} \textcolor{violet}{\textbf{A11}}} & \textcolor{violet}{\textbf{0.586}}             & \textcolor{violet}{\textbf{0.678}}          & \textbf{\textcolor{red}{$\uparrow$9.16\%}}         & \textcolor{violet}{\textbf{6/9}}           & {\color[HTML]{333333} \textcolor{violet}{\textbf{A71}}}  & \textcolor{violet}{\textbf{0.386}}             & \textcolor{violet}{\textbf{0.463}}          & \textbf{\textcolor{red}{$\uparrow$7.65\%}}         & \textcolor{violet}{\textbf{3/13}}          \\
{\color[HTML]{333333} \textcolor{violet}{\textbf{A12}}} & \textcolor{violet}{\textbf{0.456}}             & \textcolor{violet}{\textbf{0.618}}          & \textbf{\textcolor{red}{$\uparrow$16.18\%}}        & \textcolor{violet}{\textbf{4/1}}           & {\color[HTML]{333333} \textcolor{violet}{\textbf{A72}}}  & \textcolor{violet}{\textbf{0.377}}             & \textcolor{violet}{\textbf{0.492}}          & \textbf{\textcolor{red}{$\uparrow$11.48\%}}        & \textcolor{violet}{\textbf{2/2}}           \\
{\color[HTML]{333333} \textbf{A13}} & 0.871             & 0.908          & \textcolor{red}{$\uparrow$3.69\%}         & 54/41         & {\color[HTML]{333333} \textcolor{violet}{\textbf{A73}}}  & \textcolor{violet}{\textbf{0.305}}             & \textcolor{violet}{\textbf{0.398}}          & \textbf{\textcolor{red}{$\uparrow$9.28\%} }        & \textcolor{violet}{\textbf{1/8} }         \\
{\color[HTML]{333333} \textbf{A14}} & 0.975             & 0.978          & \textcolor{red}{$\uparrow$0.36\%}         & 108/94        & {\color[HTML]{333333} \textcolor{violet}{\textbf{A74}}}  & \textcolor{violet}{\textbf{0.553}}             & \textcolor{violet}{\textbf{0.621}}          & \textbf{\textcolor{red}{$\uparrow$6.84\%}}         & \textcolor{violet}{\textbf{5/18}}          \\
{\color[HTML]{333333} \textbf{A15}} & 0.964             & 0.964          & 0.00\%         & 100/106       & {\color[HTML]{333333} \textbf{A75}}  & 0.675             & 0.717          & \textcolor{red}{$\uparrow$4.22\%}         & 13/35         \\
{\color[HTML]{333333} \textbf{A16}} & 0.769             & 0.864          & \textcolor{red}{$\uparrow$9.52\%}         & 28/7          & {\color[HTML]{333333} \textbf{A76}}  & 0.665             & 0.728          & \textcolor{red}{$\uparrow$6.28\%}         & 12/22         \\
{\color[HTML]{333333} \textbf{A17}} & 0.748             & 0.774          & \textcolor{red}{$\uparrow$2.55\%}         & 24/61         & {\color[HTML]{333333} \textbf{A77}}  & 0.678             & 0.697          & \textcolor{red}{$\uparrow$1.92\%}         & 14/69         \\
{\color[HTML]{333333} \textbf{A18}} & 0.912             & 0.945          & \textcolor{red}{$\uparrow$3.30\%}         & 68/48         & {\color[HTML]{333333} \textbf{A78}}  & 0.757             & 0.818          & \textcolor{red}{$\uparrow$6.11\%}         & 27/23         \\
{\color[HTML]{333333} \textbf{A19}} & 0.912             & 0.923          & \textcolor{red}{$\uparrow$1.09\%}         & 69/82         & {\color[HTML]{333333} \textbf{A79}}  & 0.819             & 0.838          & \textcolor{red}{$\uparrow$1.91\%}         & 37/70         \\
{\color[HTML]{333333} \textbf{A20}} & 0.952             & 0.978          & \textcolor{red}{$\uparrow$2.57\%}         & 91/60         & {\color[HTML]{333333} \textbf{A80}}  & 0.977             & 0.981          & \textcolor{red}{$\uparrow$0.35\%}         & 111/99        \\
{\color[HTML]{333333} \textbf{A21}} & 0.978             & 0.967          & \textcolor{blue}{$\downarrow$-1.10\%}        & 112/117       & {\color[HTML]{333333} \textbf{A81}}  & 0.869             & 0.899          & \textcolor{red}{$\uparrow$2.97\%}         & 52/53         \\
{\color[HTML]{333333} \textbf{A22}} & 0.920             & 0.956          & \textcolor{red}{$\uparrow$3.65\%}         & 71/42         & {\color[HTML]{333333} \textbf{A82}}  & 0.654             & 0.760          & \textcolor{red}{$\uparrow$10.61\%}        & 11/4          \\
{\color[HTML]{333333} \textbf{A23}} & 0.898             & 0.934          & \textcolor{red}{$\uparrow$3.65\%}         & 61/42         & {\color[HTML]{333333} \textbf{A83}}  & 0.718             & 0.760          & \textcolor{red}{$\uparrow$4.17\%}         & 19/36         \\
{\color[HTML]{333333} \textbf{A24}} & 0.931             & 0.942          & \textcolor{red}{$\uparrow$1.09\%}         & 78/83         & {\color[HTML]{333333} \textbf{A84}}  & 0.680             & 0.727          & \textcolor{red}{$\uparrow$4.72\%}         & 15/29         \\
{\color[HTML]{333333} \textbf{A25}} & 0.807             & 0.836          & \textcolor{red}{$\uparrow$2.92\%}         & 34/56         & {\color[HTML]{333333} \textbf{A85}}  & 0.869             & 0.890          & \textcolor{red}{$\uparrow$2.09\%}         & 53/66         \\
{\color[HTML]{333333} \textbf{A26}} & 0.960             & 0.964          & \textcolor{red}{$\uparrow$0.36\%}         & 96/95         & {\color[HTML]{333333} \textbf{A86}}  & 0.737             & 0.817          & \textcolor{red}{$\uparrow$8.01\%}         & 22/12         \\
{\color[HTML]{333333} \textbf{A27}} & 0.996             & 1.000          & \textcolor{red}{$\uparrow$0.36\%}         & 120/96        & {\color[HTML]{333333} \textbf{A87}}  & 0.957             & 0.960          & \textcolor{red}{$\uparrow$0.35\%}         & 95/101        \\
{\color[HTML]{333333} \textbf{A28}} & 0.855             & 0.898          & \textcolor{red}{$\uparrow$4.36\%}         & 44/31         & {\color[HTML]{333333} \textbf{A88}}  & 0.932             & 0.964          & \textcolor{red}{$\uparrow$3.13\%}         & 79/52         \\
{\color[HTML]{333333} \textcolor{violet}{\textbf{A29}}} & \textcolor{violet}{\textbf{0.607}}             & \textcolor{violet}{\textbf{0.709}}          & \textbf{\textcolor{red}{$\uparrow$10.18\%}}        & \textcolor{violet}{\textbf{7/5}}           & {\color[HTML]{333333} \textbf{A89}}  & 0.809             & 0.802          & \textcolor{blue}{$\downarrow$-0.70\%}        & 36/115        \\
{\color[HTML]{333333} \textcolor{violet}{\textbf{A30}}} & \textcolor{violet}{\textbf{0.644}}             & \textcolor{violet}{\textbf{0.745}}          & \textbf{\textcolor{red}{$\uparrow$10.18\%}}        & \textcolor{violet}{\textbf{9/5}}           & {\color[HTML]{333333} \textbf{A90}}  & 0.856             & 0.878          & \textcolor{red}{$\uparrow$2.26\%}         & 45/64         \\
{\color[HTML]{333333} \textbf{A31}} & 0.750             & 0.804          & \textcolor{red}{$\uparrow$5.43\%}         & 26/27         & {\color[HTML]{333333} \textbf{A91}}  & 0.732             & 0.796          & \textcolor{red}{$\uparrow$6.45\%}         & 21/21         \\
{\color[HTML]{333333} \textbf{A32}} & 0.888             & 0.917          & \textcolor{red}{$\uparrow$2.90\%}         & 59/57         & {\color[HTML]{333333} \textbf{A92}}  & 0.935             & 0.949          & \textcolor{red}{$\uparrow$1.41\%}         & 83/80         \\
{\color[HTML]{333333} \textbf{A33}} & 0.902             & 0.920          & \textcolor{red}{$\uparrow$1.81\%}         & 62/72         & {\color[HTML]{333333} \textbf{A93}}  & 0.845             & 0.873          & \textcolor{red}{$\uparrow$2.78\%}         & 42/59         \\
{\color[HTML]{333333} \textbf{A34}} & 0.866             & 0.862          & \textcolor{blue}{$\downarrow$-0.36\%}        & 51/112        & {\color[HTML]{333333} \textbf{A94}}  & 0.845             & 0.850          & \textcolor{red}{$\uparrow$0.52\%}         & 41/92         \\
{\color[HTML]{333333} \textbf{A35}} & 0.975             & 0.978          & \textcolor{red}{$\uparrow$0.36\%}         & 109/97        & {\color[HTML]{333333} \textbf{A95}}  & 0.972             & 0.983          & \textcolor{red}{$\uparrow$1.05\%}         & 106/86        \\
{\color[HTML]{333333} \textbf{A36}} & 0.956             & 0.935          & \textcolor{blue}{$\downarrow$-2.18\%}        & 94/120        & {\color[HTML]{333333} \textbf{A96}}  & 0.963             & 0.972          & \textcolor{red}{$\uparrow$0.87\%}         & 99/88         \\
{\color[HTML]{333333} \textbf{A37}} & 0.830             & 0.899          & \textcolor{red}{$\uparrow$6.88\%}         & 39/17         & {\color[HTML]{333333} \textbf{A97}}  & 0.986             & 0.990          & \textcolor{red}{$\uparrow$0.35\%}         & 117/103       \\
{\color[HTML]{333333} \textbf{A38}} & 0.935             & 0.953          & \textcolor{red}{$\uparrow$1.81\%}         & 81/71         & {\color[HTML]{333333} \textbf{A98}}  & 0.988             & 0.993          & \textcolor{red}{$\uparrow$0.52\%}         & 118/93        \\
{\color[HTML]{333333} \textbf{A39}} & 0.924             & 0.957          & \textcolor{red}{$\uparrow$3.26\%}         & 76/50         & {\color[HTML]{333333} \textbf{A99}}  & 0.960             & 0.963          & \textcolor{red}{$\uparrow$0.35\%}         & 96/102        \\
{\color[HTML]{333333} \textbf{A40}} & 0.946             & 0.953          & \textcolor{red}{$\uparrow$0.72\%}         & 86/89         & {\color[HTML]{333333} \textbf{A100}} & 0.934             & 0.949          & \textcolor{red}{$\uparrow$1.57\%}         & 80/75         \\
{\color[HTML]{333333} \textbf{A41}} & 0.703             & 0.812          & \textcolor{red}{$\uparrow$10.87\%}        & 17/3          & {\color[HTML]{333333} \textbf{A101}} & 0.951             & 0.955          & \textcolor{red}{$\uparrow$0.35\%}         & 90/99         \\
{\color[HTML]{333333} \textbf{A42}} & 0.964             & 0.993          & \textcolor{red}{$\uparrow$2.90\%}         & 100/57        & {\color[HTML]{333333} \textbf{A102}} & 0.953             & 0.955          & \textcolor{red}{$\uparrow$0.17\%}         & 92/105        \\
{\color[HTML]{333333} \textbf{A43}} & 0.985             & 0.971          & \textcolor{blue}{$\downarrow$-1.45\%}        & 116/119       & {\color[HTML]{333333} \textbf{A103}} & 0.739             & 0.772          & \textcolor{red}{$\uparrow$3.30\%}         & 23/47         \\
{\color[HTML]{333333} \textbf{A44}} & 0.801             & 0.877          & \textcolor{red}{$\uparrow$7.61\%}         & 31/14         & {\color[HTML]{333333} \textbf{A104}} & 0.920             & 0.957          & \textcolor{red}{$\uparrow$3.65\%}         & 74/44         \\
{\color[HTML]{333333} \textbf{A45}} & 0.902             & 0.913          & \textcolor{red}{$\uparrow$1.09\%}         & 62/85         & {\color[HTML]{333333} \textcolor{violet}{\textbf{A105}}} & \textcolor{violet}{\textbf{0.650}}             & \textcolor{violet}{\textbf{0.711}}          & \textbf{\textcolor{red}{$\uparrow$6.09\%}}         & \textcolor{violet}{\textbf{10/24}}         \\
{\color[HTML]{333333} \textbf{A46}} & 0.902             & 0.938          & \textcolor{red}{$\uparrow$3.62\%}         & 62/46         & {\color[HTML]{333333} \textcolor{violet}{\textbf{A106}}} & \textcolor{violet}{\textbf{0.623}}             & \textcolor{violet}{\textbf{0.690}}          & \textbf{\textcolor{red}{$\uparrow$6.78\%}}         & \textcolor{violet}{\textbf{8/19}}          \\
{\color[HTML]{333333} \textbf{A47}} & 0.841             & 0.884          & \textcolor{red}{$\uparrow$4.35\%}         & 40/34         & {\color[HTML]{333333} \textbf{A107}} & 0.717             & 0.753          & \textcolor{red}{$\uparrow$3.65\%}         & 18/44         \\
{\color[HTML]{333333} \textbf{A48}} & 0.851             & 0.851          & 0.00\%         & 43/106        & {\color[HTML]{333333} \textbf{A108}} & 0.905             & 0.922          & \textcolor{red}{$\uparrow$1.74\%}         & 66/74         \\
{\color[HTML]{333333} \textbf{A49}} & 0.887             & 0.953          & \textcolor{red}{$\uparrow$6.55\%}         & 57/20         & {\color[HTML]{333333} \textbf{A109}} & 0.903             & 0.932          & \textcolor{red}{$\uparrow$2.96\%}         & 65/54         \\
{\color[HTML]{333333} \textbf{A50}} & 0.858             & 0.898          & \textcolor{red}{$\uparrow$4.01\%}         & 47/38         & {\color[HTML]{333333} \textbf{A110}} & 0.750             & 0.821          & \textcolor{red}{$\uparrow$7.13\%}         & 25/16         \\
{\color[HTML]{333333} \textbf{A51}} & 0.935             & 0.946          & \textcolor{red}{$\uparrow$1.09\%}         & 81/83         & {\color[HTML]{333333} \textbf{A111}} & 0.948             & 0.958          & \textcolor{red}{$\uparrow$1.04\%}         & 88/87         \\
{\color[HTML]{333333} \textbf{A52}} & 0.964             & 0.978          & \textcolor{red}{$\uparrow$1.45\%}         & 100/78        & {\color[HTML]{333333} \textbf{A112}} & 0.981             & 0.988          & \textcolor{red}{$\uparrow$0.69\%}         & 113/91        \\
{\color[HTML]{333333} \textbf{A53}} & 0.946             & 0.931          & \textcolor{blue}{$\downarrow$-1.45\% }       & 86/118        & {\color[HTML]{333333} \textbf{A113}} & 0.990             & 0.990          & 0.00\%         & 119/106       \\
{\color[HTML]{333333} \textbf{A54}} & 0.880             & 0.920          & \textcolor{red}{$\uparrow$3.99\%}         & 56/39         & {\color[HTML]{333333} \textbf{A114}} & 0.965             & 0.958          & \textcolor{blue}{$\downarrow$-0.69\%}        & 103/114       \\
{\color[HTML]{333333} \textbf{A55}} & 0.982             & 0.982          & 0.00\%         & 115/106       & {\color[HTML]{333333} \textbf{A115}} & 0.927             & 0.950          & \textcolor{red}{$\uparrow$2.26\%}         & 77/65         \\
{\color[HTML]{333333} \textbf{A56}} & 0.862             & 0.920          & \textcolor{red}{$\uparrow$5.80\%}         & 48/26         & {\color[HTML]{333333} \textbf{A116}} & 0.939             & 0.977          & \textcolor{red}{$\uparrow$3.82\%}         & 84/40         \\
{\color[HTML]{333333} \textbf{A57}} & 0.924             & 0.956          & \textcolor{red}{$\uparrow$3.27\%}         & 75/49         & {\color[HTML]{333333} \textbf{A117}} & 0.915             & 0.932          & \textcolor{red}{$\uparrow$1.74\%}         & 70/73         \\
{\color[HTML]{333333} \textbf{A58}} & 0.960             & 0.960          & 0.00\%         & 98/106        & {\color[HTML]{333333} \textbf{A118}} & 0.920             & 0.950          & \textcolor{red}{$\uparrow$2.96\%}         & 72/55         \\
{\color[HTML]{333333} \textbf{A59}} & 0.982             & 0.996          & \textcolor{red}{$\uparrow$1.47\%}         & 114/76        & {\color[HTML]{333333} \textbf{A119}} & 0.920             & 0.941          & \textcolor{red}{$\uparrow$2.09\%}         & 72/67         \\
{\color[HTML]{333333} \textbf{A60}} & 0.949             & 0.964          & \textcolor{red}{$\uparrow$1.45\%}         & 89/79         & {\color[HTML]{333333} \textbf{A120}} & 0.976             & 0.972          & \textcolor{blue}{$\downarrow$-0.35\%}        & 110/111  \\\hline

\multicolumn{2}{c}{\multirow{2}{*}{Average Positive Gain/Number: +3.76\%/105}}& \multicolumn{2}{c}{\multirow{2}{*}{Average Negative Gain/Number: -0.94\%/10}}& \multicolumn{3}{c}{\multirow{2}{*}{Maximum Gain of $hard$ Action: \textbf{16.18\%}}}& \multicolumn{3}{c}{\multirow{2}{*}{Average Gain of $hard$ Action: \textbf{9.38\%}}} \\
\multicolumn{10}{c}{}\\\bottomrule      
\end{tabular}}

\label{tab333}
\end{table*}

\textbf{Efficiency comparison.} \cref{tab:FLOP} compares the efficiency with various state-of-the-art methods. Compared to the temporal Transformer method ST-TR\cite{ST-TR}, our approach uses only 1/5 of its parameters and 1/31 of its FLOPs, while achieving a 5.2\% improvement in accuracy. Compared to the temporal pooling method Koopman\cite{Koopman}, our method achieves a 2.2\% performance gain with only half the number of parameters. \cref{fig:IT} compares the inference time of several advanced methods under the same hardware configuration. For a  fair comparison, we measured the average time after 300 rounds of iteration. Encouragingly, TSkel-Mamba achieves the optimal recognition accuracy with  highly competitive inference speed, which is attributed to the properties of  Mamba, also providing the inspiration for some real-time tasks  $e.g.$ online action recognition~\cite{InfoGCNplus}.

\textbf{Visual validation of TDM's effectiveness: }The \textit{t-SNE}~\cite{TSNE} algorithm is utilized to project the high-dimensional action features output from the last layer of Our method onto a 2D plane.  \cref{fig:TSNE} compares the \textit{t-SNE} features generated by $Baseline$ with and without our proposed TDM block. 6 categories were randomly selected, and different colors were assigned to them.  The features are approximately clustered into 6 components.   After the TDM block is incorporated,  the distance within each cluster is reduced,  showing a more compact and denser morphology. Consequently, TDM can contribute to the more effective alignment of features with category semantics. The effectiveness of TDM for temporal dynamic modeling is interpretively verified from a visualization perspective.

\begin{figure}[!]
	\centering 
	\includegraphics[width=0.49\textwidth]{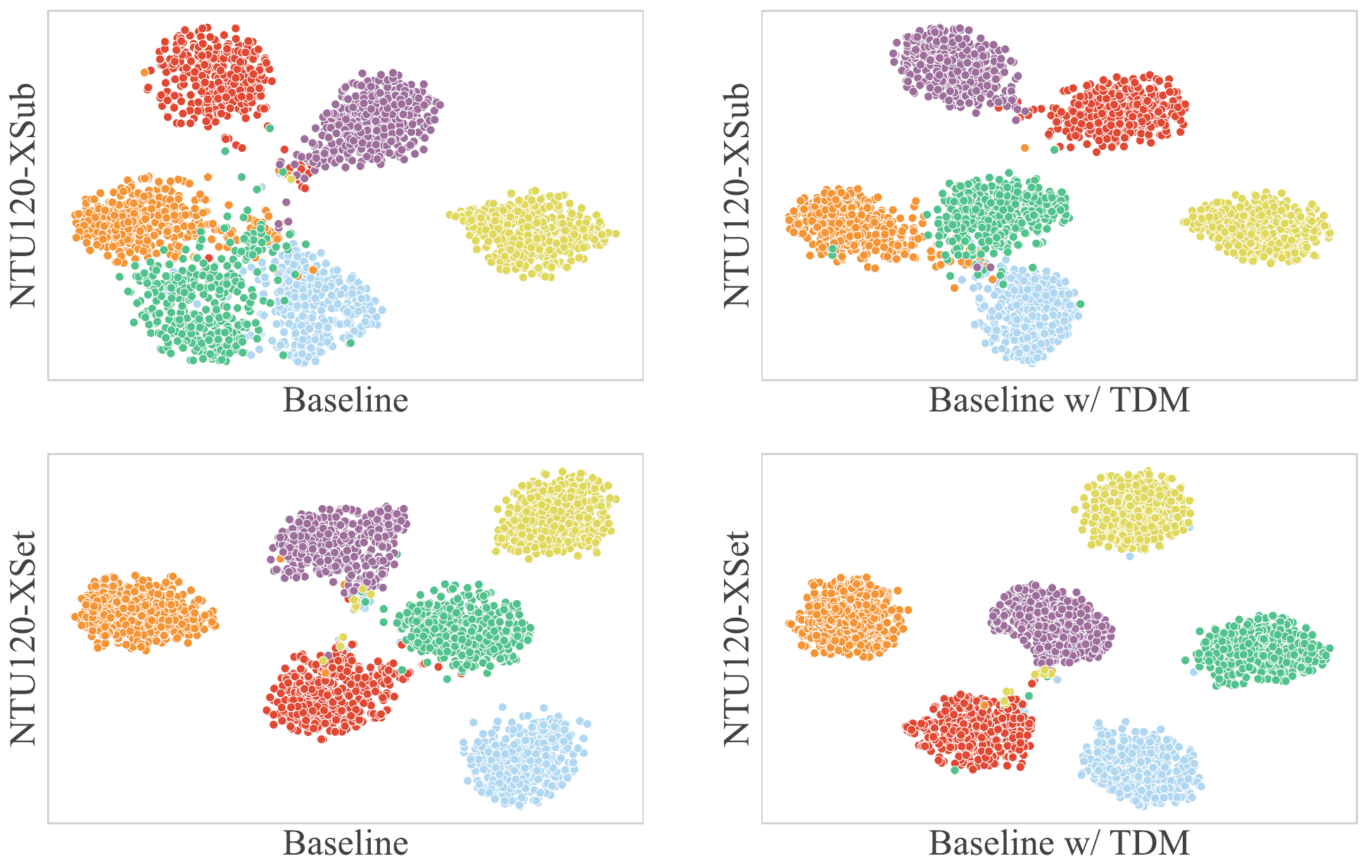}
	\caption{Comparison of \textit{t-SNE}  between the $Baseline$ with and without TDM block  under X-Set and X-Sub benchmarks on NTU120 dataset. We randomly picked 6  categories, with each represented by a different color. $w/o$ means without. }
	\label{fig:TSNE}
\end{figure}

\textbf{Bidirectional Temporal Modeling Discussion.} Our TDM framework first downsamples the channel size of the feature map and learns bidirectional temporal information in two separate streams, which are then concatenated. While this approach is not the key contribution of our work, and similar concepts have been explored in video action recognition, it remains effective. Intuitively, for bidirectional temporal modeling, we could implement an existing method where the feature map is not downsampled, and the MTI-enhanced feature map is fed into a Bi-SSM, which consists of two internal streams, and the output feature maps of the two streams are summed to generate the final output. In this baseline, we achieve a performance of 86.7\% on the X-Sub subset of NTU 120, with a model size of 3.9M parameters. In comparison, our method achieve 87.4\% with 2.4M parameters, meaning the baseline model has 1.5M more parameters and a 0.7\% lower performance. As discussed in~\cref{subsec33}, Channel Projection, our approach reduces model parameters, making the model easier to optimize. Additionally, the concatenation operation allows for more informative feature fusion from both the forward and backward streams, leading to better action recognition.

\textbf{TSkel-Mamba's Advantage.} We would like to provide some clarification for gains and highlight where the proposed TSkel-Mamba excels: 

First, while the margin of performance gains on the NTU dataset may seem modest, it is meaningful on this highly saturated benchmark—where even a 0.1\% gain corresponds to over 100 more correctly classified samples. Top-performing methods such as GAP\cite{GAP}, FR-Head\cite{FR-Head}, Koopman\cite{Koopman}, and others have also reported incremental gains in this range on NTU benchmarks, yet TSkel-Mamba further excels in efficiency and generalizability. 

Next, TSkel-Mamba introduces significantly less complexity than Koopman ( Requires 50\% fewer parameters, Has lower FLOPs; Delivers 2.2\% gain over Koopman under fair comparisons in \cref{tab:FLOP} ). This positions TSkel-Mamba as a cost-efficient solution, especially relevant for scenarios such as edge deployment, robotics, or mobile applications, where both accuracy and computational footprint are critical.

Finally,  TSkel-Mamba shows clear strengths in challenging or underexplored scenarios, where Koopman and similar methods struggle:  1) \textsl{Long-Term Action Sequences. } As shown in  \cref{fig:gains}, when frame counts increase beyond 128 (e.g., up to 256), baseline models suffer performance drops due to increased temporal complexity. In contrast, TSkel-Mamba continues to improve, achieving over 4.5\% gain at 256 frames. This underscores its robustness in modeling long-range temporal dependencies. 2) \textsl{Hard Action Recognition.} In challenging classes with subtle motion cues or intra-class similarity, our Temporal Dependency Modeling (TDM) module shows significant gains. According to \cref{tab:R55}, TDM achieves 16.18\% maximum gain and  9.38\% average gain across hard classes. These improvements demonstrate TSkel-Mamba's enhanced discriminative capability in complex settings. 3) \textsl{Cross-Dataset Generalizability.}  To further validate generalizability, we evaluated on UAV-Human, a challenging large-scale dataset with 155 action classes and diverse real-world conditions. Compared with Koopman on $\mathbb{S}_4$ metrics, TSkel-Mamba achieves up to 3.0\% performance gain with only ~1/2 the parameters. This highlights the model’s scalability and adaptability beyond NTU-style indoor datasets.

\section{Limitations}
In the design of spatotemporal backbone for skeletal data,  the prohibitive computational costs of temporal Transformers makes them suboptimal condidates for temporal modeling.   Our proposed Temporal Dynamics Modeling (TDM) block is a powerful and novel temporal plugin with lower computational overhead compared to temporal Transformers.   Although TDM demonstrates significantly stronger temporal dynamic modeling capabilities, we observe that the Mamba architecture requires slightly higher computational parameters than temporal convolutions (TCN).   Therefore, in specific architectural configurations, both pure TDM ($e.g.$, our proposed TSkel-Mamba) and hybrid TDM-TCN architectures (compatibility comparisons in \cref{tab5}) can achieve better action recognition.  Furthermore, we evaluated TDM's performance gains across varying sequence lengths in \cref{fig:gains}.  Its advantages are less pronounced when processing very short sequences, the results nevertheless reveal TDM's promising potential for long-term action understanding, warranting further exploration in future work.

\section{Conclusion}

This work introduced a novel Mamba-based temporal modeling solution—the Temporal Dynamic Modeling (TDM) block—for skeleton-based action recognition. Integrated with spatial Transformers, TDM forms the foundation of our proposed backbone, TSkel-Mamba, which achieves state-of-the-art performance while maintaining competitive inference speed. Beyond this architecture, TDM demonstrates strong compatibility when incorporated into other leading spatial-dominant models, highlighting its potential as a versatile and effective temporal plugin for skeleton data. Looking ahead, we aim to extend the application of TDM to language-supervised and online action recognition, as well as explore its utility in downstream tasks involving skeleton feature extractors. Additionally, the enhanced cross-channel temporal interactions enabled by TDM offer promising directions for other temporal tasks e.g., motion generation. Limitations and more details are discussed in the \textsl{Appendix}.





\bibliographystyle{IEEEtran}
\bibliography{BiB}
\newpage

\vfill

\end{document}